\documentclass{article}

\usepackage[submission]{colm2026_conference}

\usepackage[T1]{fontenc}
\usepackage[utf8]{inputenc}
\usepackage{newtxtext,newtxmath}
\usepackage{inconsolata}

\usepackage{kotex}

\usepackage{amsmath}
\usepackage{mathtools}
\usepackage{latexsym}

\usepackage{booktabs}
\usepackage{array}
\usepackage{tabularx}
\usepackage{multirow}
\usepackage{makecell}
\usepackage{siunitx}
\usepackage{colortbl}
\usepackage{arydshln}
\usepackage{wrapfig} 

\usepackage{graphicx}
\usepackage{float}
\usepackage{placeins}
\usepackage{cuted}

\usepackage{xcolor}
\usepackage{color}

\usepackage{tcolorbox}
\tcbuselibrary{breakable, skins}

\usepackage{caption}
\usepackage{enumitem}
\usepackage{comment}
\usepackage{soul}
\usepackage{microtype}
\usepackage{setspace}
\usepackage{lineno}
\usepackage{pifont}

\usepackage{url}
\usepackage{hyperref}
\usepackage[capitalize]{cleveref}

\hypersetup{
    colorlinks,
    linkcolor={blue},
    citecolor={blue},
    urlcolor={blue}
}

\newcommand{\algname}{\textsc{SoCRATES}}

\definecolor{absgray}{RGB}{242,243,245}
\definecolor{metablue}{RGB}{0,102,204}

\definecolor{tim}{HTML}{C0504D}
\definecolor{eff}{HTML}{6BAE5A}
\definecolor{cgn}{HTML}{4F81BD}

\newcolumntype{L}[1]{>{\raggedright\let\newline\\\arraybackslash\hspace{0pt}}m{#1}}
\newcolumntype{X}[1]{>{\centering\let\newline\\\arraybackslash\hspace{0pt}}p{#1}}

\newtcolorbox{promptbox}[1]{
    breakable,
    colback=white,
    colframe=black!70,
    fonttitle=\scriptsize\bfseries,
    title=#1,
    toprule=1.5pt,
    bottomrule=0.5pt,
    left=3pt,
    right=3pt,
    top=3pt,
    bottom=3pt,
}

\newtcolorbox{systembox}{
    breakable,
    colback=white,
    colframe=black!40,
    leftrule=3pt,
    rightrule=0.4pt,
    toprule=0.4pt,
    bottomrule=0.4pt,
    fontupper=\scriptsize,
    title={\scriptsize\textbf{System}},
    left=3pt,
    right=3pt,
    top=2pt,
    bottom=2pt,
    before skip=2pt,
    after skip=2pt,
}

\newtcolorbox{userbox}{
    breakable,
    colback=white,
    colframe=black!20,
    leftrule=3pt,
    rightrule=0.4pt,
    toprule=0.4pt,
    bottomrule=0.4pt,
    fontupper=\scriptsize,
    title={\scriptsize\textbf{User}},
    left=3pt,
    right=3pt,
    top=2pt,
    bottom=2pt,
    before skip=2pt,
    after skip=2pt,
}

\newcommand{\customabstractpage}{
\begin{tcolorbox}[
    enhanced,
    colback=absgray,
    colframe=absgray,
    boxrule=0pt,
    arc=8pt,
    left=3mm,
    right=3mm,
    top=3mm,
    bottom=3mm
]

\newcommand{\titlelogo}{\raisebox{-0.25\height}{\includegraphics[height=1.5em]{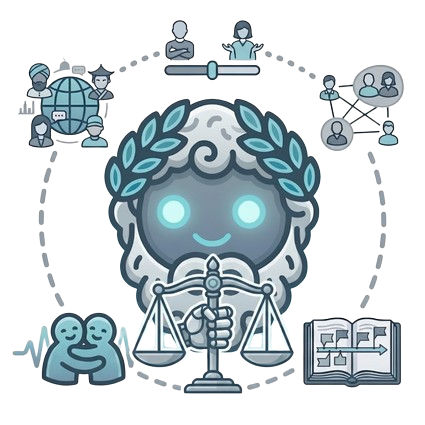}}}

{\Large\bfseries
\titlelogo \algname{}: Towards Reliable Automated Evaluation of Proactive LLM Mediation across Domains and Socio-cognitive Variations
\par}

\author{Taewon Yun, Hyeonseong Park, Jeonghwan Choi, Hayoon Park, Yeeun Choi, Hwanjun Song\thanks{Corresponding Author.}\\
Korea Advanced Institute of Science and Technology\\
\texttt{\{ytaewon0415, songhwanjun\}@kaist.ac.kr}}

\vspace{3mm}

Taewon Yun$^{1}$, Hyeonseong Park$^{1}$, Jeonghwan Choi$^{1}$, Hayoon Park$^{1}$, Yeeun Choi$^{2}$, \\
Hwanjun Song$^{1}$\par

\vspace{1mm}

$^{1}$KAIST, $^{2}$Chungnam National University\par

\vspace{4mm}

\noindent
Evaluating LLM mediators remains challenging, as mediation unfolds as a real-time trajectory shaped by disputants' shifting emotions, intentions, and context. Existing testbeds rely on a few expert-authored domains, vary mainly strategic posture, and score every turn against every topic, introducing off-topic noise. We introduce \algname{}, a benchmark for evaluating proactive LLM mediators in realistic, multi-domain testbeds. It constructs scenarios from real conflicts through an agentic pipeline across eight domains, probes five socio-cognitive adaptation axes (strategic posture, party composition, history length, emotional reactivity, and cultural identity), and scores each topic only on the turns that advance it via a topic-localized evaluator. The evaluator reaches 0.82 alignment with human experts, more than doubling a per-turn baseline. Benchmarking eight frontier LLMs, we find that even the strongest mediator closes only about a third of the unmediated consensus gap under diverse and realistic testbeds, with performance varying sharply by socio-cognitive axis, highlighting that progress lies in social adaptation to diverse conditions.
\vspace{4mm}

\noindent
\begin{minipage}[t]{0.65\textwidth}
{\small
\textbf{Date:} June 3, 2026 \par
\textbf{Correspondence:} Hwanjun Song at {\color{metablue}\href{mailto:songhwanjun@kaist.ac.kr}{songhwanjun@kaist.ac.kr}} \par
\textbf{First Author:} Taewon Yun at {\color{metablue}\href{mailto:ytaewon0415@kaist.ac.kr}{ytaewon0415@kaist.ac.kr}} \par
\textbf{Project Page:}{ \color{metablue}\url{https://disl-lab.github.io/SoCRATES}}
}
\end{minipage}
\hfill
\begin{minipage}[t]{0.27\textwidth}
\vspace*{-0.1cm}
\raggedleft
\includegraphics[width=1.0\linewidth]{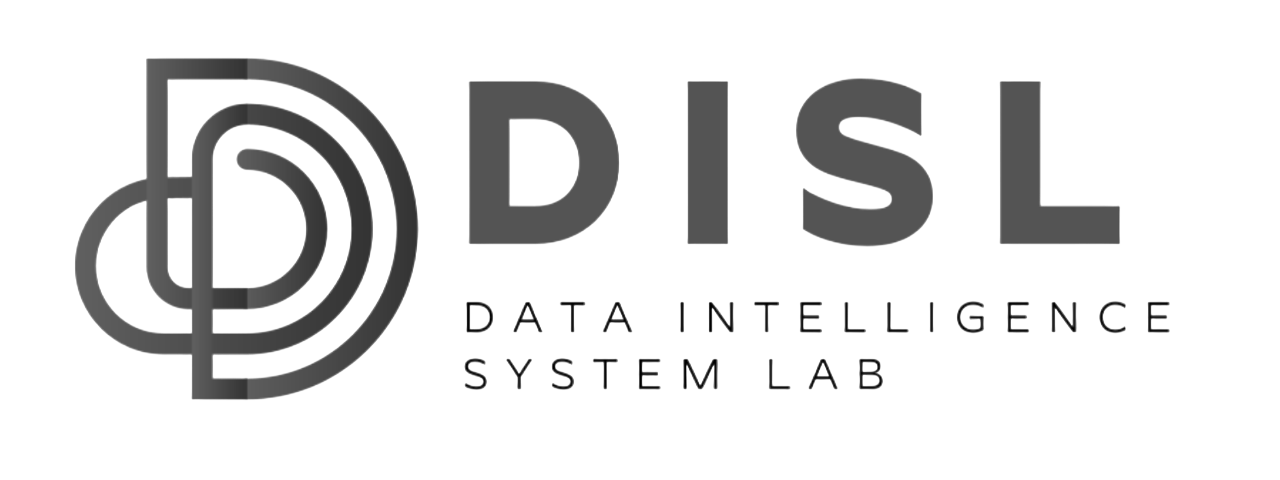}
\end{minipage}

\end{tcolorbox}
}

\begin{document}

\thispagestyle{empty}

\vspace*{-1.3cm}
\customabstractpage

\section{Introduction}
\label{sec:introduction}

Social conflict imposes heavy societal costs, yet skilled human mediators remain scarce \citep{tessler2024habermas,ma2025towards}. This has motivated efforts to deploy LLMs as automated mediators. Yet despite frontier models reaching near-expert performance on olympiad and research-level mathematics \citep{dekoninck2026matharena}, LLM mediators close only a modest fraction of the unmediated consensus gap \citep{liu2025promediate} and collapse under the variations real conflicts exhibit \citep{shapira2024clever,wu2026social}. \emph{Closing this gap is less bottlenecked by modeling than by evaluation}, since mediation has no single correct answer and must be judged on a real-time trajectory shaped by disputants' shifting emotions, intentions, and evolving context.

Building such an evaluation framework poses three challenges. First, scenario coverage does not scale, as real disputes carry privacy and legal sensitivity that confine existing testbeds to a few expert-authored domains, such as bargaining \citep{hale2025kodis} and legal disputes \citep{chen2026simulating}. Second, real-world complexity must be reproduced along multiple independent axes, since conflicts vary along disputants' emotion, culture, and history \citep{rakshit2025emotionally,guo2025conflict}, yet prior testbeds vary only strategic posture \citep{liu2025promediate,chen2026simulating}, conflating these axes and obscuring which one a mediator fails on. Third, evaluation must be both trajectory-aware and noise-resilient, since mediation quality emerges across turns rather than at an end state, yet protocols like ProMediate score every topic at every turn with an LLM judge \citep{liu2025promediate}, letting off-topic content distort scores and compound errors along the trajectory.

Prior work has advanced each challenge under inherent trade-offs \citep{tessler2024habermas,hale2025kodis,chen2026simulating,liu2025promediate}, trading realism for scalability, scalability for interactivity, or interactivity for reliability. We contend that mediation evaluation requires a unified leap, namely an automated pipeline that scales scenario coverage across diverse real conflicts, varies socio-cognitive axes independently to localize where mediators fail, and scores trajectories reliably end-to-end.

\begin{figure*}[t]
\centering
\includegraphics[height=3.5cm]{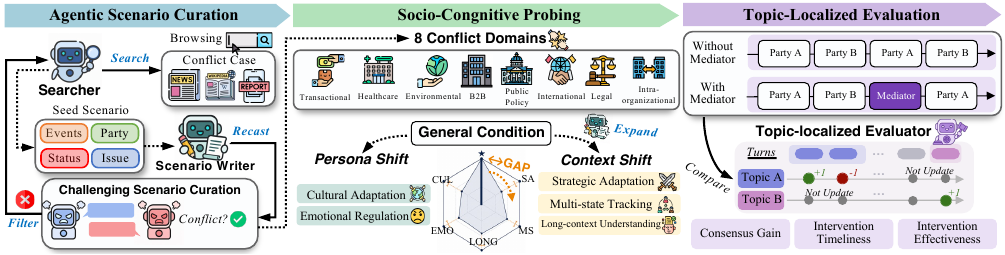}
\vspace{-0.2cm}
\caption{Overview of \algname{}: agentic scenario curation grounds scenarios in a real conflict, socio-cognitive probing expands scenarios along five axes to expose where mediators fails, and topic-localized evaluation scores each trajectory with three metrics to quantify the mediator's contribution.}
\label{fig:overview}
\vspace{-0.3cm}
\end{figure*}

We thus realize this leap in \algname{} (\underline{So}cial \underline{C}onflict \underline{R}esolution \underline{A}rena with \underline{T}opic-localized \underline{E}valuation for \underline{S}ocial Cognition), illustrated in Figure \ref{fig:overview}. \algname{} addresses the three challenges through three stages, curating real-grounded scenarios, probing them along socio-cognitive axes, and evaluating trajectories topic-by-topic.

\smallskip\smallskip
\noindent\textbf{Agentic Scenario Curation.} \algname{} treats scenario construction itself as an \emph{agentic process} that scales without human authoring. A three-stage pipeline orchestrates LLM agents that \emph{(i) search} the web for real public disputes across eight conflict domains, including transactional, healthcare, business, and legal, \emph{(ii) recast} each retrieved case into a structured scenario, and \emph{(iii) filter} the pool through rejection sampling, retaining only hard scenarios that fail to resolve in unmediated simulation.

\smallskip\smallskip
\noindent\textbf{Socio-Cognitive Probing.} \algname{} uses these curated scenarios as the simulation testbed and probes mediator behavior across five socio-cognitive axes, restructured from prior literature on mediator competencies \citep{susskind1999consensus,bowling2000bringing,lebaron2003bridging} to expose where each mediator fails. We vary \emph{(i) strategic posture} (e.g., competing vs. accommodating) to probe strategic adaptation, \emph{(ii) party composition} (two- vs. three-disputant) to probe multi-state tracking, \emph{(iii) history length} (short vs. extended background) to probe long-context understanding, \emph{(iv) emotional reactivity} (composed vs. reactive) to probe emotional regulation, and \emph{(v) cultural identity} (different cultural profiles) to probe cultural adaptation. As axes are applied independently rather than stacked, any shift in mediator performance is attributable to a single axis, yielding per-axis diagnostics of mediator competence.

\smallskip\smallskip
\noindent\textbf{Topic-Localized Evaluation.} To enable real-time, multi-faceted scoring of mediator trajectories, \algname{} introduces a \emph{topic-localized} evaluator that, for each topic, scores agreement only at the turns that actively move it and carries scores forward otherwise. The evaluator supports three complementary metrics: \emph{(i) consensus gain} measures the mediator's overall contribution to closing the unmediated agreement gap, \emph{(ii) intervention timeliness} measures when the mediator acts relative to escalation, and \emph{(iii) intervention effectiveness} measures how much each intervention shifts consensus. Validated against two expert annotators on 1,844 dialogue snippets, our evaluator correlates with experts at a Pearson coefficient of $0.82$ on the trajectory and $0.80$ at the outcome level, more than doubling both ProMediate's per-turn evaluator \citep{liu2025promediate} and a non-expert baseline.

Our main contributions are: (1) \algname{}, a unified, automated evaluation framework for proactive LLM mediation that integrates agentic scenario curation, socio-cognitive probing, and topic-localized evaluation in a single pipeline; (2) a topic-localized evaluator that scores mediator trajectories along three real-time metrics and exhibits high correlation with expert judgments; (3) a comprehensive benchmark of eight proprietary and open-source LLM mediators across diverse conflict domains and socio-cognitive axes; and (4)  we find that the strongest mediator closes only roughly a third of the unmediated consensus gap under diverse and realistic testbeds, and that gains vary sharply by socio-cognitive axis, with strong mediation adapting its intervention strategy to socio-cognitive demands.

\section{Related Work}
\label{sec:relatedwork}

\noindent\textbf{{Social Conflict Resolution.}}
Social conflict resolution steers disputing parties toward a consensus, dynamically intervening as the dispute unfolds to defuse it before escalation~\citep{deutsch2011handbook}. Prior work frames this as negotiation, casting LLMs as the disputing parties to study bargaining~\citep{bianchi2024well, zhou2024sotopia, kwon2024llms}. While this direction shows that they can faithfully reproduce human social behavior in conflicts, it does not reveal how disputes between humans are resolved.
Beyond this, recent work positions the LLM as the third-party mediator that, given a complete recorded conversation, finds common ground and proposes a solution~\citep{tan2024robots, tessler2024habermas}. Yet such studies must recruit thousands of human disputants to simulate conflicts~\citep{chawla2021casino, hale2025kodis, tessler2024habermas}, posing a scalability bottleneck. 
Thus, building on evidence that LLMs reproduce the behavior of disputants, recent work utilizes LLM simulation to enable scalable testbeds~\citep{chen2026simulating}. Among these, Promediate~\citep{liu2025promediate} proposes a proactive agent that decides when and how to intervene at the interaction level, better matching the dynamic intervention conflict resolution demands. As mediation shifts to the interaction level, it increasingly calls for capabilities long emphasized in social reasoning and multi-turn interaction, such as adapting to parties that differ in mental state~\citep{xiao2025towards} and cultural background~\citep{ki2025multiple}, and to varied context~\citep{shapira2024clever}.


\smallskip\smallskip\noindent\textbf{Automated Dialogue Evaluation.} 
Evaluating multi-turn dialogues through human judgment is costly and difficult to scale~\citep{zheng2023judging, deshpande2025multichallenge}. This motivates the adoption of automatic evaluators for interaction assessment. 
Specifically, in negotiation and mediation, such approaches judge dialogue progress indirectly through end-state outcomes such as consensus or goal achievement between parties~\citep{zhou2024sotopia, chen2026simulating}.
Yet this single signal alone provides only a coarse view of dialogue state, and recent work shows that decomposing evaluation into fine-grained, turn-level signals across topics yields a more faithful, trajectory-level representation of how the conversation unfolds~\citep{mannekote2023agreement, zhang2025sotopia, liu2025promediate}.
Tracking every topic, however, remains difficult. LLM judges have long been known to treat unrelated content as noise that distracts their judgment~\citep{ye2025justice}, and in trajectory evaluation such errors propagate to subsequent states~\citep{liu2025promediate}. Reducing this noise has thus become an increasingly important direction for reliable dialogue
evaluation.




\section{\algname{} Framework}
\label{sec:method}

We formalize the mediation task and then build \algname{} in three stages: agentic scenario curation assembles the scenario pool from real public disputes, socio-cognitive probing expands each scenario along five axes, and topic-localized evaluation scores every trajectory with three metrics.

\subsection{Task Formulation}
\label{sec:method:task}

Following the widely adopted Harvard conflict simulation framework~\citep{fisher2011getting}, we cast social conflict as the negotiation of a fixed topic set by parties with divergent positions, and represent a conflict scenario as a tuple $s = (\mathcal{B}, \mathcal{P}, \mathcal{T}, \mathcal{W})$.
The background $\mathcal{B}$ collects the past histories, prior commitments, and strategic posture of the conflict, which together form the common ground from which every party reasons. The party set $\mathcal{P} = \{p_1, \dots, p_n\}$ denotes the disputants, with $n \geq 2$. The topic set $\mathcal{T} = \{T_1, \dots, T_k\}$ enumerates the points of conflict, where each topic carries a discrete option set, making movement observable as a shift among options rather than free-form text. The preference set $\mathcal{W} = \{w_1, \dots, w_n\}$ assigns each party a weight vector $w_i$ over the topics summing to $100$, encoding how much each topic matters and keeping the disagreement non-trivial yet resolvable.

\smallskip\smallskip\noindent\textbf{{Disputing Parties.}}
Within a scenario, each party $p_i$ is an LLM agent that speaks on its turn, conditioned on two inputs. The shared input, visible to all, is the background $\mathcal{B}$, the topics $\mathcal{T}$, and the dialogue so far. The private input, visible only to $p_i$, is its profile: an objective, a fallback if talks fail, and a per-topic starting stance, a persona $\pi_i$ setting its emotional and cultural identity, and preferences $\mathcal{W}$. \algname{}'s socio-cognitive conditions perturb one scenario component at a time, either a party's profile, the background $\mathcal{B}$, or the party set $\mathcal{P}$ (\S\ref{sec:method:simulation}).

\smallskip\smallskip\noindent\textbf{{Mediator.}}
A third-party mediator observes the exchange and may speak between party turns. Unlike a party, it sees only the shared input, the background $\mathcal{B}$, the topics $\mathcal{T}$, and the dialogue so far, never any party's persona, stance, or preferences. Thus, the mediator must infer these hidden states from the dialogue, making mediation a test of social cognition. Each turn, it decides when to intervene and, if so, how to move the parties toward agreement across the topics. \algname{} scores both the when and the how of each intervention within the mediation.

\subsection{Agentic Scenario Curation}
\label{sec:method:construction}

Prior testbeds rely on human experts who hand-crafted scenarios from commercial resources~\citep{liu2025promediate} or government databases~\citep{chen2026simulating}, capping coverage at the few domains these experts can reach. We instead curate every scenario from a real conflict via agentic deep research, where LLMs retrieve and synthesize web evidence across domains while staying faithful to cited sources~\citep{gou2026mind2web, tao2025webshaper}. \algname{} chains this into a three-step pipeline: a \textit{Searcher} gathers real conflict cases, a \textit{Scenario Writer} recasts them into enactable scenarios, and an unmediated simulation filters out cases that resolve on their own.

\smallskip\smallskip\noindent\textbf{{Seed Scenario Search.}}
We span eight domains (transactional, healthcare, environmental, business-to-business, public-policy, international, legal, and intra-organizational), each a canonical class of disputes drawn from Harvard teaching materials. For each domain, a \textit{Searcher} agent (o4-mini-deep-research~\citep{openai2025deepresearch}) takes the domain as a query and gathers conflict cases from the web, compiling each dispute's parties, contested topics, and event history into a \emph{seed}.

\smallskip\smallskip\noindent\textbf{{Scenario Recast.}}
A raw seed is in report form and cannot be enacted directly, so a \textit{Scenario Writer} agent (GPT-5.4, chosen for its strong long-form writing ability) recasts each seed into the structured scenario of \S\ref{sec:method:task}, comprising a background $\mathcal{B}$, a party set $\mathcal{P}$ with roles and per-topic stances, a topic set $\mathcal{T}$ with options, and a preference allocation $\mathcal{W}$ for each party, conditioned on the seed's background, topics, and party profiles. Prompts and example scenarios are provided in Appendix~\ref{app:scenario_detail}, with Table~\ref{tab:scenario-example} confirming that recast scenarios faithfully preserve their source inspirations.

\smallskip\smallskip\noindent\textbf{{Simulation-based Filtering.}}
A mediator can only be credited for resolving a conflict that would not have resolved on its own, so we keep only scenarios that fail to resolve unmediated. Each candidate is enacted as a multi-turn dialogue, the \emph{general} simulation that also serves as the unperturbed baseline for later expansion. Parties are role-playing agents (DeepSeek-V3.2)\footnotemark[1] held fixed across all runs, taking turns in a fixed cyclic order and emitting a private inner thought with each utterance to stay consistent with their role and persona~\citep{liu2025proactive, liu2025promediate}. As LLMs lack a natural stopping point and would otherwise talk past agreement or loop indefinitely~\citep{hu2026multi}, we adopt a explicit termination criteria. Simulations end as resolved once every party signals consensus, or as an impasse when a party walks away or the $100$-turn budget is reached.

\footnotetext[1]{We select DeepSeek-V3.2 for its ability to faithfully reproduce assigned personas (see \S\ref{sec:validation:sim}).}

We run this simulation three times per candidate without a mediator and retain the scenario only when all three replays end in impasse. Rejected scenarios feed back to the \textit{Searcher} for a fresh seed, until \algname{} accumulates $40$ hard scenarios, five per domain, forming the general condition.

\subsection{Socio-Cognitive Probing}
\label{sec:method:simulation}

Starting from the $40$ general-condition scenarios, \algname{} probes mediator behavior along five socio-cognitive axes, running every resulting condition both with and without a mediator.

\smallskip\smallskip\noindent\textbf{{Socio-cognitive Condition Expansion.}}
The five axes are restructured from core mediator competencies~\citep{susskind1999consensus,bowling2000bringing,lebaron2003bridging} and organized into two groups, a \emph{context} group that raises the cognitive load of the conflict itself and a \emph{persona} group that varies disputant identity. Although stacking axes would reflect real-world complexity, it would entangle failures across competencies and obscure the performance gap attributable to each social variation. We therefore apply each axis independently to a fresh copy of the scenario, so any change in mediator performance traces back to one competency. 

The context group comprises three axes that perturb the conflict itself:
\begin{itemize}[leftmargin=*, noitemsep, topsep=2pt]
\item \emph{Strategic Posture} specifies one of three Thomas-Kilmann conflict modes~\citep{thomas2008thomas} in the background $\mathcal{B}$, \emph{competing} (prioritizing self-interest), \emph{avoiding} (withdrawing from conflict), or \emph{accommodating} (placing others' interests ahead of one's own), to probe strategic adaptation.
\item \emph{Party Composition} adds a third disputant, synthesized by the \textit{Scenario Writer} from the scenario again, to probe multi-state tracking.
\item \emph{History Length} has the \textit{Scenario Writer} expand the past histories and prior commitments within the background to five times its default length, to probe long-context understanding.
\end{itemize}

\footnotetext[2]{We adopt Hofstede's cultural values~\citep{hofstede2010cultures} (\emph{e.g.}, uncertainty avoidance, individualism) as cultural background, since they shape conflict-handling~\citep{caputo2019relationship} and underlie surface customs and religion~\citep{guo2025conflict}.}

The persona group comprises two axes that vary disputant identity, each applied by adding a persona instruction to the party profile. The two axes are:
\begin{itemize}[leftmargin=*, noitemsep, topsep=2pt]
\item \emph{Emotional Reactivity} sets each party's reactivity on a $0$--$1$ scale (higher = more reactive), fixed at the two endpoints, composed (Com, $0$) and reactive (React, $1$), to keep the contrast sharp, yielding three unordered party pairings.
\item \emph{Cultural Identity} anchors each party to a Korean (KR), American (US), or Chinese (CN) identity through Hofstede profiles to probe cultural adaptation.\footnotemark[2] Identities are encoded as a statement summarizing its 0--100 scores across the six Hofstede dimensions, appended to the party profile. To isolate identity from language, we prompt all parties to interact in English. This yields three intra-cultural and three cross-cultural pairings.
\end{itemize}

Together with the general condition, the five axes yield 15 conditions. Refer to Appendix~\ref{app:probing} for the full list of conditions with their prompts and effects on conflict dynamics.

\subsection{Topic-Localized Evaluation}
\label{sec:method:evaluation}

\subsubsection{Benchmark Metrics}
\label{sec:method:evaluation:metrics}

\algname{} compares each mediator against the matched unmediated run to quantify added consensus. For each topic $T_j \in \mathcal{T}$, the evaluator outputs a 1--5 agreement rating, which we remap to $[0,1]$ and average across topics into a \emph{Consensus Score} $S_{\leq t} \in [0,1]$ at every turn $t$. Here, $S_{\leq t}$ snapshots the cumulative consensus state up to turn $t$, rather than the agreement at turn $t$ alone, enabling two of our metrics to track real-time dynamics rather than only terminal outcomes. Each scenario therefore yields two matched trajectories, $\{S^\mathrm{unmed}_{\leq t}\}$ and $\{S^\mathrm{med}_{\leq t}\}$, on which the three metrics below operate.

\smallskip\smallskip\noindent\textbf{{Intervention Timeliness.}}
This metric captures \emph{when} a mediator acts, rewarding a prompt response once consensus drops within the mediated trajectory. We call a turn $t_\mathrm{drop}$ a \emph{drop event} when $S^\mathrm{med}_{\leq t}$ falls by at least $\tau = 0.1$ relative to the preceding turn, and let $t_\mathrm{s}$ be the first intervention within the next $W = 10$ turns:
\[
\text{Intervention Timeliness} =
\left(1 - \frac{t_\mathrm{s} - t_\mathrm{drop}}{W}\right) \times 100,
\]
averaged across drop events in a run, where $100$ corresponds to an immediate response and $0$ to no intervention within the window.

\smallskip\smallskip\noindent\textbf{{Intervention Effectiveness.}}
This metric captures \emph{how} effective each mediator utterance is, the consensus lift it produces over the following five turns. For an intervention at turn $i$,
\[\text{Intervention Effectiveness} = \frac{S^\mathrm{med}_{\leq i+5} - S^\mathrm{med}_{\leq i-1}}{1 - S^\mathrm{med}_{\leq i-1}} \times 100,\]
averaged across a mediator's interventions, where $S^\mathrm{med}_{\leq i-1}$ and $S^\mathrm{med}_{\leq i+5}$ are the consensus snapshots immediately before and five turns after the utterance. The normalization by $1 - S^\mathrm{med}_{\leq i-1}$ accounts for ceiling effects when consensus is already high, while negative values indicate interventions that reduce consensus.

\smallskip\smallskip\noindent\textbf{{Consensus Gain.}} This metric measures a mediator's overall contribution as the fraction of the unmediated consensus gap closed at the end state.
\[\text{Consensus Gain} =
\frac{S^\mathrm{med} - S^\mathrm{unmed}}{1 - S^\mathrm{unmed}} \times 100,\]
where $S^\mathrm{unmed}$ and $S^\mathrm{med}$ are the terminal Consensus Scores of the matched runs without and with a mediator. Normalizing by the remaining gap $1 - S^\mathrm{unmed}$ makes scenarios with different initial states comparable. A value of $100$ closes the gap entirely, while a negative value indicates the parties end up worse off than without a mediator. When $S^\mathrm{unmed} = 1$, we report the raw change $S^\mathrm{med} - S^\mathrm{unmed}$.

\subsubsection{Automatic Evaluation}
\label{sec:method:evaluation:auto}

Per-turn LLM judges score every topic at every turn~\citep{liu2025promediate}, yet only a few topics are actively contested at any given turn while the rest stay inactive, so scoring inactive topics injects noise from irrelevant content~\citep{koo2024benchmarking, ye2025justice} and compounds errors along the trajectory. We instead localize scoring to the turns that move each topic. For each topic $T_j$, the judge reads the dialogue once and locates the turns where $T_j$ is actively in play, those at which it is discussed or a party shifts position. At each located turn it records an agreement score and each party's current stance, and turns that do not touch $T_j$ inherit the prior score. The full trajectory is thus recovered in a single judge pass after the conversation ends, with DeepSeek-V3.2 as the backbone. Automatic evaluation prompts in Appendix~\ref{app:evaluator_detail}.

We validate this evaluator against expert raters in \S\ref{sec:validation:trace}, where it reaches a Pearson $r = 0.82$ with experts, more than doubling both \textsc{ProMediate}~\citep{liu2025promediate} and a non-expert baseline.

\begin{table*}[t]
\centering
\renewcommand{\arraystretch}{1.1}
\footnotesize
\setlength{\tabcolsep}{5pt}
\begin{tabular}{|l|c|c|c|c|c|c|c|c|}
\specialrule{\heavyrulewidth}{0pt}{0pt}
\makecell{Simulator} & \makecell{DeepSeek\\-V3.2} & \makecell{Gemini-3.1\\-Pro} & \makecell{GPT-5.4} & \makecell{Gemini-3.1\\-FL} & \makecell{Qwen3\\-235B} & \makecell{GPT-5.4\\-mini} & \makecell{Qwen3\\-30B} & \makecell{Kripp.'s $\alpha$\\(IAA)} \\
\specialrule{\lightrulewidth}{0pt}{0pt}
Persona & \textbf{87.2} & 86.9 & 80.4 & 75.0 & 74.7 & 72.5 & 70.4 & 0.75 \\
\specialrule{\heavyrulewidth}{0pt}{0pt}
\end{tabular}
\vspace{-0.2cm}
\caption{Simulation fidelity for persona fidelity (accuracy (\%) via A/B comparison based evaluation)}
\label{tab:persona-reactiveness}
\vspace{-0.5cm}
\end{table*}

\section{Validation of \algname{}}
\label{sec:validation}
Two components of \algname{} require empirical validation before benchmarking: (i) the disputant simulators must actually produce the prescribed persona variations\footnotemark[3], and (ii) the topic-localized evaluator must trace trajectories reliably. We validate (i) by checking whether the persona scalar steers party behavior as intended, and (ii) via alignment with human expert judgments.

\footnotetext[3]{We focus on persona fidelity because, for remaining axes, the perturbations are structural by construction or supported by prior validations of strategic posture~\citep{liu2025promediate} and cultural persona~\citep{dey2025can}.}

\paragraph{Simulation Fidelity.}
\label{sec:validation:sim}
\algname{} uses a float-valued intensity scalar when expanding each party persona, and we ask whether varying this scalar steers agent behavior. We operationalize the check through reactiveness, the persona dimension governing emotional escalation. To probe intensity controllability beyond the binary, we test four scalar levels $\{0, 0.33, 0.66, 1\}$ and check whether agents preserve this scale across simulated conversations.

We evaluate seven strong simulators, drawn from the mediator pool and supplemented with updated backbones (GPT-5.4, Gemini-3.1-Pro), to isolate persona controllability from weak simulator failure. Following the protocol of \citet{choi2026makes}, we sample two levels at random from the four-level grid, pair each against a third randomly chosen reference, and generate the conversations with reference's persona held fixed. Human annotators select the more reactive side in each pair, and higher annotator accuracy indicates more faithful intensity control (see Appendix~\ref{app:simulation_validation_detail} for annotation details).

This yields 160 A/B pairs per simulator, annotated by three crowdworkers with Krippendorff's $\alpha = 0.75$. DeepSeek-V3.2 achieves the highest score (Table~\ref{tab:persona-reactiveness}), indicating that the float-valued persona reliably translates into ordered reactiveness.

\begin{wraptable}{r!}{0.5\textwidth} 
\vspace{-0.4cm}
\centering
\renewcommand{\arraystretch}{1.1}
\footnotesize
\begin{tabular}{|L{1.7cm}|X{2cm}|X{1.9cm}|}
\specialrule{\heavyrulewidth}{1pt}{0pt}
Evaluator & Trajectory level & Outcome level \\
\specialrule{\lightrulewidth}{0pt}{0pt}
\textsc{Non-expert} & 0.331 (0.000) & 0.527 (0.000) \\
\textsc{ProMediate}        & 0.372 (0.000) & 0.432 (0.000) \\
\algname{}     & \textbf{0.823} (0.000) & \textbf{0.801} (0.000) \\
\specialrule{\heavyrulewidth}{0pt}{1pt}
\end{tabular}
\vspace{-0.1cm}
\caption{Evaluator alignment with experts (Pearson $r$). The values in parenthesis represent p-values.}
\label{tab:trace-validation}
\vspace{-0.4cm}
\end{wraptable}

\paragraph{Topic-localized Evaluation.}
\label{sec:validation:trace}
We test whether the topic-localized evaluation tracks expert judgment. Since humans recognize consensus only after a claim has been met with a response~\citep{clark1991grounding}, per-turn human annotation would inject ambiguity. We instead aggregate the evaluator's per-turn trajectory into snippets, single back-and-forth exchanges, and have experts annotate at this unit. Aggregation preserves any per-turn evaluator error while matching the resolution at which experts can rate reliably. Two expert annotators rate 1,844 snippets from 144 mediator trajectories, sampled to ensure balanced coverage across domains and  models under the same 1--5 rubric as the evaluator (see Appendix~\ref{app:evaluation_validation_detail} for annotation details), reaching inter-annotator agreement of $\alpha = 0.86$.

We compare \algname{} against two baselines on the same 1--5 scale: \textsc{ProMediate}'s LLM judge, which scores every topic at every turn regardless of relevance, and a \textsc{Non-expert} rater performing the same task as the experts. We measure alignment with the average expert score using Pearson correlation $r$ at two granularities: \emph{trajectory-level} (all snippets) and \emph{outcome-level} (the final snippet). The two views complement each other, as intervention quality metrics depend on conflict trajectory while consensus gain depends on the final state.

The topic-localized evaluator achieves the strongest alignment with experts, reaching $r = 0.82$ on trajectories and $r = 0.80$ on outcomes (Table~\ref{tab:trace-validation}), more than doubling both baselines on trajectories. This result remains consistent under another backbone (see Table~\ref{tab:trace-validation-other} in Appendix~\ref{app:evaluation_validation_detail}). Without localization, per-turn baselines distort the consensus trajectory (see Figure~\ref{fig:trend} in Appendix~\ref{app:evaluation_validation_detail}).

\begin{table*}[t]
\scriptsize
\centering
\renewcommand{\arraystretch}{1.15}
\setlength{\tabcolsep}{2pt}
\resizebox{\textwidth}{!}{%
\begin{tabular}{@{}c@{\hspace{2pt}}l@{\hspace{2pt}} |ccccccccc | ccccccccc |ccccccccc@{}}
\toprule
\multirow{2}{*}{Type} & \multirow{2}{*}{Mediator}
 & \multicolumn{9}{c}{\textbf{Intervention Timeliness}}
 & \multicolumn{9}{c}{\textbf{Intervention Effectiveness}}
 & \multicolumn{9}{c}{\textbf{Consensus Gain}} \\
\cmidrule(lr){3-11} \cmidrule(lr){12-20} \cmidrule(lr){21-29}
 &
 & Trans & Heal & Env & B2B & Pol & Intl & Legal & Intra & \textit{Avg.}
 & Trans & Heal & Env & B2B & Pol & Intl & Legal & Intra & \textit{Avg.}
 & Trans & Heal & Env & B2B & Pol & Intl & Legal & Intra & \textit{Avg.} \\
\specialrule{\lightrulewidth}{1pt}{0pt}
\multirow{2}{*}{\rotatebox[origin=c]{90}{\textbf{Prop.}}}
  & Gemini-3.1-FL & \cellcolor{tim!65} 81.2 & \cellcolor{tim!65} 84.1 & \cellcolor{tim!55} 78.2 & \cellcolor{tim!65} 81.8 & \cellcolor{tim!65} 82.9 & \cellcolor{tim!65} 81.4 & \cellcolor{tim!5} 72.9 & \cellcolor{tim!65} 84.4 & \cellcolor{tim!65} 80.9 & \cellcolor{eff!65} 33.6 & \cellcolor{eff!95} \textbf{27.8} & \cellcolor{eff!55} 16.7 & \cellcolor{eff!95} \textbf{23.5} & \cellcolor{eff!80} 30.3 & \cellcolor{eff!55} 19.5 & \cellcolor{eff!65} 29.4 & \cellcolor{eff!80} 16.1 & \cellcolor{eff!80} 24.6 & \cellcolor{cgn!65} 52.1 & \cellcolor{cgn!80} 47.7 & \cellcolor{cgn!55} 25.9 & \cellcolor{cgn!95} \textbf{34.6} & \cellcolor{cgn!95} \textbf{36.0} & \cellcolor{cgn!55} 22.0 & \cellcolor{cgn!65} 26.7 & \cellcolor{cgn!65} 18.8 & \cellcolor{cgn!80} 33.0 \\
  & GPT-5.4-mini & \cellcolor{tim!55} 80.7 & \cellcolor{tim!55} 81.9 & \cellcolor{tim!65} 82.3 & \cellcolor{tim!45} 76.3 & \cellcolor{tim!45} 78.2 & \cellcolor{tim!45} 77.2 & \cellcolor{tim!65} 78.6 & \cellcolor{tim!55} 84.3 & \cellcolor{tim!55} 79.9 & \cellcolor{eff!95} \textbf{34.9} & \cellcolor{eff!5} 18.9 & \cellcolor{eff!95} \textbf{24.6} & \cellcolor{eff!80} 23.3 & \cellcolor{eff!55} 22.5 & \cellcolor{eff!65} 21.2 & \cellcolor{eff!95} \textbf{32.3} & \cellcolor{eff!95} \textbf{18.8} & \cellcolor{eff!65} 24.6 & \cellcolor{cgn!95} \textbf{55.6} & \cellcolor{cgn!20} 23.6 & \cellcolor{cgn!95} \textbf{35.0} & \cellcolor{cgn!80} 32.0 & \cellcolor{cgn!55} 28.2 & \cellcolor{cgn!80} 30.3 & \cellcolor{cgn!95} \textbf{41.2} & \cellcolor{cgn!95} \textbf{29.5} & \cellcolor{cgn!95} \textbf{34.4} \\
\specialrule{\lightrulewidth}{0pt}{0pt}
\multirow{6}{*}{\rotatebox[origin=c]{90}{\textbf{Open-Source}}}
  & DeepSeek-V3.2 & \cellcolor{tim!35} 76.1 & \cellcolor{tim!20} 76.6 & \cellcolor{tim!35} 77.1 & \cellcolor{tim!20} 74.6 & \cellcolor{tim!20} 76.8 & \cellcolor{tim!35} 75.2 & \cellcolor{tim!45} 76.3 & \cellcolor{tim!5} 73.8 & \cellcolor{tim!20} 75.8 & \cellcolor{eff!55} 32.1 & \cellcolor{eff!20} 19.4 & \cellcolor{eff!65} 17.3 & \cellcolor{eff!65} 22.2 & \cellcolor{eff!65} 28.0 & \cellcolor{eff!80} 21.6 & \cellcolor{eff!80} 30.4 & \cellcolor{eff!35} 13.8 & \cellcolor{eff!55} 23.1 & \cellcolor{cgn!80} 53.3 & \cellcolor{cgn!65} 41.2 & \cellcolor{cgn!80} 27.6 & \cellcolor{cgn!45} 26.4 & \cellcolor{cgn!80} 35.4 & \cellcolor{cgn!65} 26.6 & \cellcolor{cgn!80} 27.0 & \cellcolor{cgn!55} 17.8 & \cellcolor{cgn!65} 31.9 \\
  & Qwen3-235B & \cellcolor{tim!20} 71.7 & \cellcolor{tim!35} 79.7 & \cellcolor{tim!45} 77.1 & \cellcolor{tim!35} 76.1 & \cellcolor{tim!35} 77.2 & \cellcolor{tim!20} 73.5 & \cellcolor{tim!55} 77.1 & \cellcolor{tim!35} 78.6 & \cellcolor{tim!35} 76.4 & \cellcolor{eff!80} 34.0 & \cellcolor{eff!55} 24.2 & \cellcolor{eff!35} 15.2 & \cellcolor{eff!55} 22.1 & \cellcolor{eff!95} \textbf{31.5} & \cellcolor{eff!95} \textbf{25.9} & \cellcolor{eff!55} 28.0 & \cellcolor{eff!65} 16.0 & \cellcolor{eff!95} \textbf{24.6} & \cellcolor{cgn!55} 51.0 & \cellcolor{cgn!35} 29.7 & \cellcolor{cgn!20} 22.8 & \cellcolor{cgn!55} 28.2 & \cellcolor{cgn!65} 32.5 & \cellcolor{cgn!95} \textbf{33.8} & \cellcolor{cgn!45} 20.7 & \cellcolor{cgn!80} 26.9 & \cellcolor{cgn!55} 30.7 \\
  & Nemotron-3-120B & \cellcolor{tim!5} 70.1 & \cellcolor{tim!5} 70.7 & \cellcolor{tim!5} 74.1 & \cellcolor{tim!5} 69.3 & \cellcolor{tim!5} 71.5 & \cellcolor{tim!5} 70.9 & \cellcolor{tim!20} 73.6 & \cellcolor{tim!20} 75.7 & \cellcolor{tim!5} 72.0 & \cellcolor{eff!35} 29.4 & \cellcolor{eff!65} 25.2 & \cellcolor{eff!5} 11.3 & \cellcolor{eff!45} 19.1 & \cellcolor{eff!35} 17.2 & \cellcolor{eff!45} 18.5 & \cellcolor{eff!20} 17.7 & \cellcolor{eff!55} 15.4 & \cellcolor{eff!35} 19.2 & \cellcolor{cgn!35} 41.9 & \cellcolor{cgn!55} 41.1 & \cellcolor{cgn!5} 16.7 & \cellcolor{cgn!5} 14.5 & \cellcolor{cgn!35} 15.8 & \cellcolor{cgn!35} 17.7 & \cellcolor{cgn!35} 7.0 & \cellcolor{cgn!20} 8.3 & \cellcolor{cgn!35} 20.4 \\
  & Solar-Pro-3 & \cellcolor{tim!80} 83.0 & \cellcolor{tim!95} \textbf{86.9} & \cellcolor{tim!95} \textbf{84.4} & \cellcolor{tim!80} 84.5 & \cellcolor{tim!80} 85.0 & \cellcolor{tim!95} \textbf{82.4} & \cellcolor{tim!95} \textbf{85.2} & \cellcolor{tim!80} 85.9 & \cellcolor{tim!95} \textbf{84.6} & \cellcolor{eff!20} 24.5 & \cellcolor{eff!45} 21.8 & \cellcolor{eff!20} 13.2 & \cellcolor{eff!35} 17.8 & \cellcolor{eff!20} 15.9 & \cellcolor{eff!5} 14.4 & \cellcolor{eff!5} 16.7 & \cellcolor{eff!5} 9.1 & \cellcolor{eff!5} 16.7 & \cellcolor{cgn!20} 41.8 & \cellcolor{cgn!45} 30.1 & \cellcolor{cgn!35} 24.3 & \cellcolor{cgn!65} 28.3 & \cellcolor{cgn!5} 6.6 & \cellcolor{cgn!5} 13.4 & \cellcolor{cgn!20} 6.0 & \cellcolor{cgn!35} 8.7 & \cellcolor{cgn!20} 19.9 \\
  & Gemma-4-26B & \cellcolor{tim!45} 79.9 & \cellcolor{tim!45} 81.5 & \cellcolor{tim!20} 74.2 & \cellcolor{tim!55} 79.1 & \cellcolor{tim!55} 81.6 & \cellcolor{tim!55} 81.3 & \cellcolor{tim!35} 74.9 & \cellcolor{tim!45} 79.5 & \cellcolor{tim!45} 79.0 & \cellcolor{eff!45} 29.8 & \cellcolor{eff!35} 20.5 & \cellcolor{eff!45} 16.0 & \cellcolor{eff!5} 12.3 & \cellcolor{eff!5} 14.3 & \cellcolor{eff!20} 17.1 & \cellcolor{eff!45} 25.3 & \cellcolor{eff!20} 9.4 & \cellcolor{eff!20} 18.1 & \cellcolor{cgn!45} 42.9 & \cellcolor{cgn!5} 22.9 & \cellcolor{cgn!45} 24.6 & \cellcolor{cgn!20} 15.8 & \cellcolor{cgn!20} 7.1 & \cellcolor{cgn!20} 15.9 & \cellcolor{cgn!55} 24.4 & \cellcolor{cgn!45} 14.6 & \cellcolor{cgn!45} 21.0 \\
  & Qwen3-30B & \cellcolor{tim!95} \textbf{84.2} & \cellcolor{tim!80} 85.2 & \cellcolor{tim!80} 84.3 & \cellcolor{tim!95} \textbf{85.6} & \cellcolor{tim!95} \textbf{85.1} & \cellcolor{tim!80} 82.2 & \cellcolor{tim!80} 83.9 & \cellcolor{tim!95} \textbf{86.4} & \cellcolor{tim!80} 84.6 & \cellcolor{eff!5} 19.1 & \cellcolor{eff!80} 26.9 & \cellcolor{eff!80} 18.8 & \cellcolor{eff!20} 17.6 & \cellcolor{eff!45} 18.6 & \cellcolor{eff!35} 17.7 & \cellcolor{eff!35} 24.4 & \cellcolor{eff!45} 14.5 & \cellcolor{eff!45} 19.7 & \cellcolor{cgn!5} -7.9 & \cellcolor{cgn!95} \textbf{48.6} & \cellcolor{cgn!65} 26.3 & \cellcolor{cgn!35} 16.0 & \cellcolor{cgn!45} 17.9 & \cellcolor{cgn!45} 18.1 & \cellcolor{cgn!5} -1.2 & \cellcolor{cgn!5} 8.2 & \cellcolor{cgn!5} 15.7 \\
\specialrule{\lightrulewidth}{0pt}{1.5pt}
\multicolumn{2}{c|}{Average} & 78.4 & 80.8 & 79.0 & 78.4 & 79.8 & 78.0 & 77.8 & 81.1 & \textit{79.2} & 29.7 & 23.1 & 16.6 & 19.7 & 22.3 & 19.5 & 25.5 & 14.1 & \textit{21.3} & 41.3 & 35.6 & 25.4 & 24.5 & 22.4 & 22.2 & 19.0 & 16.6 & \textit{25.9} \\
\specialrule{\lightrulewidth}{1.5pt}{1pt}
\end{tabular}%
}
\vspace{-0.2cm}
\caption{Conflict resolution performance of the eight mediators across eight domains: Trans (Transactional), Heal (Healthcare), Env (Environmental), B2B (Business-to-Business), Pol (Public-Policy), Intl (International), Legal (Legal), and Intra (Intra-organizational). Cell color intensity increases within each column to indicate higher scores.}
\label{tab:combined_overall_domain}
\vspace{-0.5cm}
\end{table*}

\section{Benchmarking LLM Mediators}
\label{sec:experiments}

We benchmark eight LLM mediators with \algname{}: two proprietary models, GPT-5.4-mini and Gemini-3.1-Flash-Lite, and six open-source models, Gemma-4-26B-A4B-it, Qwen3-30B-Instruct, Solar-Pro-3, Nemotron3-120B-A12B, DeepSeek-V3.2, and Qwen3-235B-Instruct. This set spans two axes, proprietary versus open-source and large versus small. At every party turn, the mediator outputs a binary decision whether to intervene. When it does, it inserts a single utterance before the next party speaks; otherwise the dialogue proceeds uninterrupted. This loop repeats until termination.
Each mediator runs once on every scenario-condition pair, $40$ scenarios $\times$ $15$ conditions $= 600$ runs per mediator and $4{,}800$ in total, each paired with its no-mediator baseline. See Appendix~\ref{app:mediator_detail} for the mediator prompts.


\subsection{Performance by Conflict Domain}
\label{sec:results:domain}

\noindent\underline{\textit{Overview.}} Table~\ref{tab:combined_overall_domain} reports three metrics per mediator across eight domains. \emph{Social conflict resolution remains challenging for every benchmarked LLM, including proprietary frontier models.} Average consensus gain caps at $34.4$, splitting mediators into a top tier ($30.7$--$34.4$) and bottom tier ($15.7$--$21.0$). The split holds across domains, where means range from 41.3 to 16.6 and none clears half the unmediated gap. This gap reflects the domain diversity and social adaptation demands of \algname{}, in sharp contrast to prior works showing a resolution rate of 80--90\% in unconditioned, single domain settings~\citep{kwon2025evaluating, chen2026simulating}.

\smallskip\smallskip\noindent\textbf{Proprietary leads, scale alone does not.} The two proprietary mediators achieve higher consensus gain than the strongest open-source by 1.1--2.5 points and lead in six of eight domains. This gap persists even as open-sources close gaps on reasoning benchmarks such as AIME25~\citep{dekoninck2026matharena}. Within a family, scale helps. Qwen3-235B nearly doubles Qwen3-30B's gain. Across families, however, scale does not order the field. Nemotron-3-120B trails the smaller Gemma4-26B on Legal and Intra-organizational despite comparable problem-solving~\citep{chandiramani2026nemotron}. Together, these results show that \emph{general capability does not directly translate to mediation, and the residual gap depends strongly on the conflict domain.}

\smallskip\smallskip\noindent\textbf{Timeliness without effectiveness.} Solar-Pro-3 and Qwen3-30B post the highest intervention timeliness yet rank low on consensus gain. They intervene too often without meaningfully affecting the outcome (see Appendix~\ref{app:intervention_analysis}). Intervention effectiveness, in contrast, aligns with consensus gain. The three mediators tied at 24.6 hold the top three consensus gain scores. \textit{A good mediator must intervene at the right moments and with the right content, as timeliness alone does not resolve conflict.}

\smallskip\smallskip\noindent\textbf{{Domain coverage shapes the verdict.}}
Intervention timeliness is stable across the eight domains, whereas consensus gain swings from $41.3$ on Transactional conflict down to $16.6$ on Intra-organizational disputes. The easy end coincides with where prior conflict resolution datasets concentrate, since Transactional conflict corpora dominate existing testbeds such as CaSiNo~\citep{chawla2021casino}, CraigslistBargain~\citep{he2018decoupling}, and KODIS~\citep{hale2025kodis}. \emph{A benchmark restricted to transactional conflict overstates mediation ability, making it essential to evaluate how mediators adapt across diverse conflict domains.}

\begin{figure*}[t]
\centering
\includegraphics[width=\textwidth]{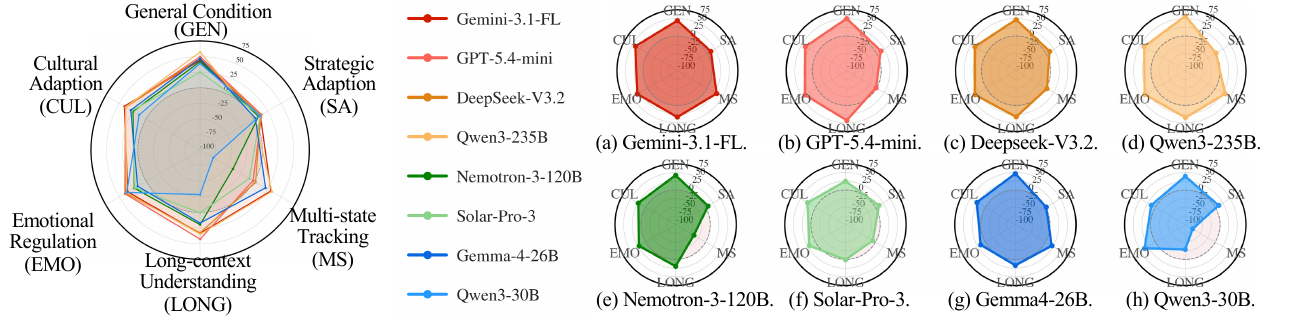}
\vspace*{-0.2cm}
\caption{Mediator adaptation across general condition and five socio-cognitive axes, measured by consensus gain.}
\label{fig:radar}
\vspace*{-0.2cm}
\end{figure*}

\subsection{Socio-cognitive Adaptation Analysis}
\label{sec:experiments:socio}

We use the five independently perturbed socio-cognitive axes to localize which abilities constrain each mediator. Figure~\ref{fig:radar} profiles each mediator across the general condition and the five axes. 

\smallskip\smallskip\noindent\underline{\textit{Highlight.}} On four of the five axes, area grows with model capability, with the proprietary models and Qwen3-235B enclosing the largest regions, yet every mediator contracts on at least one axis. Even within the top tier with comparable overall consensus gain, GPT-5.4-mini and DeepSeek-V3.2 lose far more under Multi-state Tracking than Gemini-3.1-FL and Qwen3-235B. \emph{Mediation competence therefore comprises distinct socio-cognitive abilities, and current LLMs exhibit uneven profiles rather than a single capability frontier.}


\begin{figure*}[t]
\centering
\includegraphics[width=\textwidth]{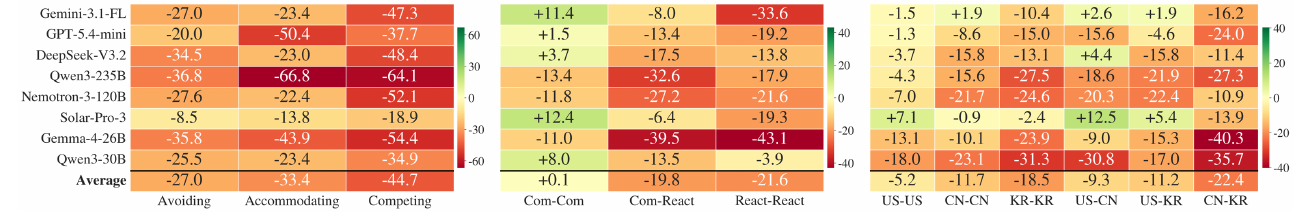}
\vspace*{-0.4cm}
{\small \hspace*{0.5cm} (a) Strategy-wise Analysis. \hspace*{1cm} (b) Emotion-wise Analysis. \hspace*{1.4cm} (c) Culture-wise Analysis.}
\vspace*{0.3cm}
\caption{Consensus gain shift from the general (unperturbed) condition along three axes: (a) strategic posture, (b) emotional reactivity, and (c) cultural identity. Negative values indicate degradation, positive values improvement.}
\label{fig:shift}
\vspace*{-0.4cm}
\end{figure*}

\subsubsection{Strategy, Emotion, and Culture Shifts}
\label{sec:experiments:delta}

The uneven model profiles motivate a closer analysis of axes. We therefore measure how consensus gain shifts from the general (unperturbed) condition when strategic posture, emotional reactivity, or cultural identity varies, as summarized in Figure~\ref{fig:shift}.

\smallskip\smallskip\noindent\textbf{Strategy.}
Strategic posture is the sharpest stress test. All non-collaborative postures reduce consensus gain, with the most severe drops under Competing ($18.9$--$64.1$) and Accommodating ($13.8$--$66.8$). Qwen3-235B suffers the largest drops in both settings despite its high overall ranking, indicating that adversarial or one-sided conflicts demand a capability that aggregate scoring does not capture.

\smallskip\smallskip\noindent\textbf{Emotion.}
Emotional reactivity produces a smoother but consistent degradation. When both parties are composed, several mediators hold their general score. When both are reactive, every mediator drops. The magnitude does not follow model size, indicating that absorbing emotional volatility, rather than raw scale, separates mediators on this axis.

\smallskip\smallskip\noindent\textbf{Culture.} Cultural identity produces the smallest but most systematic shifts, with mediator scores declining as cultural distance from U.S. norms grows. From a Hofstede perspective, all LLM mediators appear robust on U.S.-anchored values but weaker on East Asian ones, where collectivist orientation and power distance shape the dynamics differently.

\begin{figure*}[t]
\centering
\includegraphics[width=\textwidth]{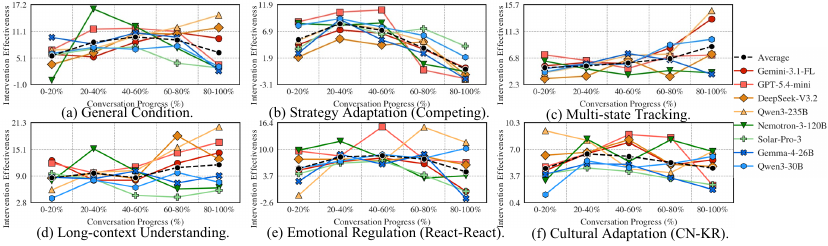}
\vspace*{-0.5cm}
\caption{Intervention Effectiveness over conversation progress, where turns are mapped to a 0--100\% scale to align varying turn counts, across the general condition and each hard condition from five socio-cognitive axes.}
\label{fig:position}
\vspace*{-0.4cm}
\end{figure*}

\subsubsection{Intervention Timing Adaptation}
\label{sec:timing}
Axis-level results show how much consensus changes, but not when interventions help. We thus analyze timing in Figure~\ref{fig:position}, which plots intervention effectiveness over normalized conversation progress for the general condition and each hard socio-cognitive condition. Since intervention effectiveness ranges differ across conditions, we read each panel as a within-condition timing profile.

The best intervention window moves with the condition. For Strategy Adaptation or Emotional Regulation, effectiveness rises early and falls off, since mediators must reframe stances or cool emotion before they harden. For Multi-state Tracking or Long-context Understanding, effectiveness instead grows toward later turns, when complex contexts make late moves like summarization more useful.
Across mediators, the key distinction is whether they follow these timing windows. Stronger mediators peak near each condition's window—GPT-5.4-mini in Strategy and Emotion, Qwen3-235B in Multi-state and Long-context—while weaker ones trace flatter curves, failing to adapt their timing as the conflict evolves. \emph{Effective mediation thus requires adapting timing to the socio-cognitive demands faced in conflict to maximize impact.}

\section{Conclusion}
\label{sec:conclusion}

We presented \algname{}, a benchmark probing LLM mediators along eight domains and five socio-cognitive axes, built on automatic scenario construction and a topic-localized evaluator. \algname{} shows that conflict resolution remains challenging for LLMs and that performance shifts across context and party compositions. This indicates that effective mediation hinges on adaptation, not uniformity, and \algname{} provides the testbed to study it.

\section*{Limitations}

While \algname{} provides a controlled testbed for evaluating LLM mediation across domains and socio-cognitive conditions, several limitations remain. First, the benchmark currently runs all conversations in English, even when parties are assigned different cultural identities. This design isolates cultural values from language variation and keeps simulator behavior comparable across conditions, but it does not test multilingual mediation. Extending \algname{} to multilingual settings would reveal how language choice, translation ambiguity, and language-specific politeness norms affect mediator behavior.

Second, \algname{} focuses on consensus as the primary outcome, since consensus is directly tied to whether a settlement is reached and can be scored consistently across domains. However, mediation quality also involves party satisfaction~\citep{hale2025kodis}, procedural fairness, trust restoration, and emotional repair. These dimensions depend on subjective party perceptions and are therefore harder to validate reliably, but incorporating well-calibrated rubrics for them would provide a more comprehensive evaluation of LLM mediators. We leave these extensions as future work.

\section*{Ethical Considerations}

We design \algname{} as a simulation study in which LLM agents role-play conflicts, so no real people are involved as disputants in this process. The scenarios are synthesized by LLM agents from deep-research seeds, with any residual references to specific individuals, organizations, or locations anonymized by the agents before the scenarios enter the benchmark.
We recruit crowd-sourced annotators and supervised graduate annotators solely for evaluator validation and persona-fidelity verification, and no other human subjects participate in social conflict simulations. Crowd-sourced annotators receive compensation above the U.S. federal minimum wage
rate, while expert examiners were compensated at
rates exceeding \$35 per hour.

\clearpage
\bibliographystyle{assets/plainnat}
\bibliography{colm2026_conference}

@article{tessler2024habermas,
  title={AI can help humans find common ground in democratic deliberation},
  author={Tessler, Michael Henry and Bakker, Michiel A. and Jarrett, Daniel and Sheahan, Hannah and Chadwick, Martin J. and Koster, Raphael and Evans, Georgina and Campbell-Gillingham, Lucy and Collins, Tantum and Parkes, David C. and Botvinick, Matthew and Summerfield, Christopher},
  journal={Science},
  year={2024}
}

@inproceedings{ma2025towards,
  title={Towards human-ai deliberation: Design and evaluation of llm-empowered deliberative ai for ai-assisted decision-making},
  author={Ma, Shuai and Chen, Qiaoyi and Wang, Xinru and Zheng, Chengbo and Peng, Zhenhui and Yin, Ming and Ma, Xiaojuan},
  booktitle={CHI},
  year={2025}
}

@inproceedings{chen2026simulating,
  title={Simulating Dispute Mediation with LLM-Based Agents for Legal Research},
  author={Chen, Junjie and Li, Haitao and Qin, Minghao and Zhou, Yujia and Ren, Yanxue and Wang, Wuyue and Liu, Yiqun and Wu, Yueyue and Ai, Qingyao},
  booktitle={AAAI},
  year={2026}
}

@article{liu2025promediate,
  title={ProMediate: A Socio-cognitive framework for evaluating proactive agents in multi-party negotiation},
  author={Liu, Ziyi and Sarrafzadeh, Bahar and Zhou, Pei and Yang, Longqi and Zhao, Jieyu and Sharma, Ashish},
  journal={arXiv preprint arXiv:2510.25224},
  year={2025}
}

@inproceedings{kwon2025evaluating,
  title={Evaluating Behavioral Alignment in Conflict Dialogue: A Multi-Dimensional Comparison of LLM Agents and Humans},
  author={Kwon, Deuksin and Shrestha, Kaleen and Han, Bin and Lee, Elena Hayoung and Lucas, Gale},
  booktitle={EMNLP},
  year={2025}
}

@article{tan2024robots,
  title={Robots in the middle: Evaluating llms in dispute resolution},
  author={Tan, Jinzhe and Westermann, Hannes and Pottanigari, Nikhil Reddy and {\v{S}}avelka, Jarom{\'\i}r and Mee{\`u}s, S{\'e}bastien and Godet, Mia and Benyekhlef, Karim},
  journal={arXiv preprint arXiv:2410.07053},
  year={2024}
}

@article{bowling2000bringing,
  title={Bringing peace into the room: The personal qualities of the mediator and their impact on the mediation},
  author={Bowling, Daniel and Hoffman, David},
  journal={Negotiation Journal},
  year={2000},
}

@inproceedings{hu2026multi,
  title={Multi-agent debate for LLM judges with adaptive stability detection},
  author={Hu, Tianyu and Tan, Zhen and Wang, Song and Qu, Huaizhi and Chen, Tianlong},
  booktitle={NeurIPS},
  year={2026}
}

@article{lebaron2003bridging,
  title={Bridging cultural conflicts: A new approach for a changing world},
  author={LeBaron, Michelle},
  year={2003},
  publisher={San Francisco: Jossey-Bass, 2003}
}

@article{thomas2008thomas,
  title={Thomas-kilmann conflict mode},
  author={Thomas, Kenneth W},
  journal={TKI Profile and Interpretive Report},
  year={2008}
}

@book{susskind1999consensus,
  title={The consensus building handbook: A comprehensive guide to reaching agreement},
  author={Susskind, Lawrence E and McKearnen, Sarah and Thomas-Lamar, Jennifer},
  year={1999},
  publisher={Sage publications}
}

@article{dekoninck2026matharena,
  title={Beyond Benchmarks: MathArena as an Evaluation Platform for Mathematics with LLMs},
  author={Dekoninck, Jasper and Jovanovi{\'c}, Nikola and Gehrunger, Tim and R{\"o}gnvalddson, K{\'a}ri and Petrov, Ivo and Sun, Chenhao and Vechev, Martin},
  journal={arXiv preprint arXiv:2605.00674},
  year={2026}
}

@article{chandiramani2026nemotron,
  title={Nemotron 3 Super: Open, Efficient Mixture-of-Experts Hybrid Mamba-Transformer Model for Agentic Reasoning},
  author={Chandiramani, Aakshita and Blakeman, Aaron and Olaoye, Abdullahi and Gupta, Abhibha and Somasamudramath, Abhilash and Khattar, Abhinav and Adesoba, Adeola and Renduchintala, Adi and Asif, Adil and Agrawal, Aditya and others},
  journal={arXiv preprint arXiv:2604.12374},
  year={2026}
}

@book{hofstede2010cultures,
  title={Cultures and Organizations: Software of the Mind},
  author={Hofstede, Geert and Hofstede, Gert Jan and Minkov, Michael},
  year={2010},
  publisher={McGraw-Hill Professional},
}

@misc{openai2025deepresearch,
  title        = {Introducing deep research},
  author       = {OpenAI},
  year         = {2025},
  howpublished = {\url{https://openai.com/index/introducing-deep-research/}},
}

@inproceedings{shapira2024clever,
  title={Clever hans or neural theory of mind? stress testing social reasoning in large language models},
  author={Shapira, Natalie and Levy, Mosh and Alavi, Seyed Hossein and Zhou, Xuhui and Choi, Yejin and Goldberg, Yoav and Sap, Maarten and Shwartz, Vered},
  booktitle={EACL},
  year={2024}
}

@article{wu2026social,
  title={Social-R1: Towards Human-like Social Reasoning in LLMs},
  author={Wu, Jincenzi and Lei, Yuxuan and Lian, Jianxun and Huang, Yitian and Zhou, Lexin and Li, Haotian and Xie, Xing and Meng, Helen},
  journal={arXiv preprint arXiv:2603.09249},
  year={2026}
}

@article{guo2025conflict,
  title={Conflict resolution in intercultural communication: strategies for managing cultural conflicts},
  author={Guo, Weihong},
  journal={Humanities and Social Sciences Communications},
  year={2025},
}

@article{rakshit2025emotionally,
  title={Emotionally-Aware Agents for Dispute Resolution},
  author={Rakshit, Sushrita and Hale, James and Chawla, Kushal and Brett, Jeanne M and Gratch, Jonathan},
  journal={arXiv preprint arXiv:2509.04465},
  year={2025}
}

@inproceedings{ye2025justice,
  title={Justice or prejudice? quantifying biases in llm-as-a-judge},
  author={Ye, Jiayi and Wang, Yanbo and Huang, Yue and Chen, Dongping and Zhang, Qihui and Moniz, Nuno and Gao, Tian and Geyer, Werner and Huang, Chao and Chen, Pin-Yu and others},
  booktitle={ICLR},
  year={2025}
}

@inproceedings{kwon2024llms,
  title={Are llms effective negotiators? systematic evaluation of the multifaceted capabilities of llms in negotiation dialogues},
  author={Kwon, Deuksin and Weiss, Emily and Kulshrestha, Tara and Chawla, Kushal and Lucas, Gale and Gratch, Jonathan},
  booktitle={Findings of EMNLP},
  year={2024}
}

@book{deutsch2011handbook,
  title={The handbook of conflict resolution: Theory and practice},
  author={Deutsch, Morton and Coleman, Peter T and Marcus, Eric C},
  year={2011},
  publisher={John Wiley \& Sons}
}

@inproceedings{xiao2025towards,
  title={Towards dynamic theory of mind: Evaluating llm adaptation to temporal evolution of human states},
  author={Xiao, Yang and Wang, Jiashuo and Xu, Qiancheng and Song, Changhe and Xu, Chunpu and Cheng, Yi and Li, Wenjie and Liu, Pengfei},
  booktitle={ACL},
  year={2025}
}

@inproceedings{ki2025multiple,
  title={Multiple LLM agents debate for equitable cultural alignment},
  author={Ki, Dayeon and Rudinger, Rachel and Zhou, Tianyi and Carpuat, Marine},
  booktitle={ACL},
  year={2025}
}

@inproceedings{zheng2023judging,
  title={Judging llm-as-a-judge with mt-bench and chatbot arena},
  author={Zheng, Lianmin and Chiang, Wei-Lin and Sheng, Ying and Zhuang, Siyuan and Wu, Zhanghao and Zhuang, Yonghao and Lin, Zi and Li, Zhuohan and Li, Dacheng and Xing, Eric and others},
  booktitle={NeurIPS},
  year={2023}
}

@inproceedings{deshpande2025multichallenge,
  title={Multichallenge: A realistic multi-turn conversation evaluation benchmark challenging to frontier llms},
  author={Deshpande, Kaustubh and Sirdeshmukh, Ved and Mols, Johannes Baptist and Jin, Lifeng and Hernandez-Cardona, Ed-Yeremai and Lee, Dean and Kritz, Jeremy and Primack, Willow E and Yue, Summer and Xing, Chen},
  booktitle={Findings of ACL},
  year={2025}
}

@article{mannekote2023agreement,
  title={Agreement tracking for multi-issue negotiation dialogues},
  author={Mannekote, Amogh and Dorr, Bonnie J and Boyer, Kristy Elizabeth},
  journal={arXiv preprint arXiv:2307.06524},
  year={2023}
}

@inproceedings{zhang2025sotopia,
  title={SOTOPIA-Ω: Dynamic Strategy Injection Learning and Social Instruction Following Evaluation for Social Agents},
  author={Zhang, Wenyuan and Liu, Tianyun and Song, Mengxiao and Li, Xiaodong and Liu, Tingwen},
  booktitle={ACL},
  year={2025}
}

@inproceedings{gou2026mind2web,
  title={Mind2web 2: Evaluating agentic search with agent-as-a-judge},
  author={Gou, Boyu and Huang, Zanming and Ning, Yuting and Gu, Yu and Lin, Michael and Qi, Weijian and Kopanev, Andrei and Yu, Botao and Jimenez Gutierrez, Bernal and Shu, Yiheng and others},
  booktitle={NeurIPS},
  year={2026}
}

@inproceedings{tao2025webshaper,
  title={Webshaper: Agentically data synthesizing via information-seeking formalization},
  author={Tao, Zhengwei and Wu, Jialong and Yin, Wenbiao and Zhang, Junkai and Li, Baixuan and Shen, Haiyang and Li, Kuan and Zhang, Liwen and Wang, Xinyu and Jiang, Yong and others},
  booktitle={ICLR},
  year={2025}
}

@inproceedings{liu2025proactive,
  title={Proactive conversational agents with inner thoughts},
  author={Liu, Xingyu Bruce and Fang, Shitao and Shi, Weiyan and Wu, Chien-Sheng and Igarashi, Takeo and Chen, Xiang'Anthony'},
  booktitle={CHI},
  year={2025}
}

@article{caputo2019relationship,
  title={The relationship between cultural values, cultural intelligence and negotiation styles},
  author={Caputo, Andrea and Ayoko, Oluremi B and Amoo, Nii and Menke, Charlott},
  journal={Journal of Business Research},
  year={2019},
}

@book{fisher2011getting,
  title={Getting to yes: Negotiating agreement without giving in},
  author={Fisher, Roger and Ury, William L and Patton, Bruce},
  year={2011},
  publisher={Penguin}
}

@inproceedings{koo2024benchmarking,
  title={Benchmarking cognitive biases in large language models as evaluators},
  author={Koo, Ryan and Lee, Minhwa and Raheja, Vipul and Park, Jong Inn and Kim, Zae Myung and Kang, Dongyeop},
  booktitle={Findings of ACL},
  year={2024}
}

@inproceedings{he2018decoupling,
  title={Decoupling strategy and generation in negotiation dialogues},
  author={He, He and Chen, Derek and Balakrishnan, Anusha and Liang, Percy},
  booktitle={EMNLP},
  year={2018}
}

@article{choi2026makes,
  title={What Makes a Sale? Rethinking End-to-End Seller--Buyer Retail Dynamics with LLM Agents},
  author={Choi, Jeonghwan and Hwang, Jibin and Sun, Gyeonghun and Ban, Minjeong and Yun, Taewon and Cheon, Hyeonjae and Song, Hwanjun},
  journal={arXiv preprint arXiv:2604.04468},
  year={2026}
}

@inproceedings{chawla2021casino,
  title={Casino: A corpus of campsite negotiation dialogues for automatic negotiation systems},
  author={Chawla, Kushal and Ramirez, Jaysa and Clever, Rene and Lucas, Gale and May, Jonathan and Gratch, Jonathan},
  booktitle={NAACL},
  year={2021}
}

@inproceedings{bianchi2024well,
  title={How well can llms negotiate? negotiationarena platform and analysis},
  author={Bianchi, Federico and Chia, Patrick John and Yuksekgonul, Mert and Tagliabue, Jacopo and Jurafsky, Dan and Zou, James},
  booktitle={ICML},
  year={2024}
}

@inproceedings{dey2025can,
  title={Can LLMs express personality across cultures? Introducing culturalpersonas for evaluating trait alignment},
  author={Dey, Priyanka and Khanter, Yugal and Bothra, Aayush and Zhao, Jieyu and Ferrara, Emilio},
  booktitle={Findings of EMNLP},
  year={2025}
}

@article{vaccaro2025advancing,
  title={Advancing AI Negotiations: A Large-Scale Autonomous Negotiation Competition},
  author={Vaccaro, Michelle and Caosun, Michael and Ju, Harang and Aral, Sinan and Curhan, Jared R},
  journal={arXiv preprint arXiv:2503.06416},
  year={2025}
}

@inproceedings{hale2025kodis,
  title={Kodis: A multicultural dispute resolution dialogue corpus},
  author={Hale, James Anthony and Rakshit, Sushrita and Chawla, Kushal and Brett, Jeanne M and Gratch, Jonathan},
  booktitle={NAACL},
  year={2025}
}

@inproceedings{zhou2024sotopia,
  title={Sotopia: Interactive evaluation for social intelligence in language agents},
  author={Zhou, Xuhui and Zhu, Hao and Mathur, Leena and Zhang, Ruohong and Yu, Haofei and Qi, Zhengyang and Morency, Louis-Philippe and Bisk, Yonatan and Fried, Daniel and Neubig, Graham and others},
  booktitle={ICLR},
  year={2024}
}

@inproceedings{clark1991grounding,
  title={Grounding in communication},
  author={Herbert H. Clark and Susan Brennan},
  booktitle={Perspectives on socially shared cognition},
  year={1991},
}

@article{openai2026gpt54,
  title={GPT-5.4 Thinking System Card},
  author={{OpenAI}},
  year={2026},
  month={March},
}

@article{google2026gemma4,
  title        = {Gemma 4 Model Card},
  author       = {{Google DeepMind}},
  year         = {2026},
  month         = {April}
}

@misc{solarpro3_2026,
  title        = {Solar Pro 3: Better Reasoning at Production Scale},
  author       = {{Upstage AI}},
  year         = {2026},
  howpublished = {\url{https://www.upstage.ai/blog/en/solar-pro-3-0127}},
}

@article{google2026geminipro,
  title={Gemini 3.1 Pro Model Card},
  author={{Google DeepMind}},
  year={2026},
  month={Feb},
}

@article{google2026geminiflash,
  title={Gemini 3.1 Flash Lite Model Card},
  author={{Google DeepMind}},
  year={2026},
  month={March},
}

@article{openai2025gpt5,
  title={GPT-5 System Card},
  author={{OpenAI}},
  year={2025},
  month={August},
}

@article{liu2025deepseek,
  title={Deepseek-v3. 2: Pushing the frontier of open large language models},
  author={Liu, Aixin and Mei, Aoxue and Lin, Bangcai and Xue, Bing and Wang, Bingxuan and Xu, Bingzheng and Wu, Bochao and Zhang, Bowei and Lin, Chaofan and Dong, Chen and others},
  journal={arXiv preprint arXiv:2512.02556},
  year={2025}
}

@article{yang2025qwen3,
  title={Qwen3 Technical Report},
  author={Yang, An and others},
  journal={arXiv preprint arXiv:2505.09388},
  year={2025}
}

\clearpage
\appendix

\begin{table}[t]
\scriptsize
\centering
\vspace{-0.4cm}
\renewcommand{\arraystretch}{1.15}
\setlength{\tabcolsep}{4pt}
\begin{tabular}{|l|l|l|l|l|}
\specialrule{\heavyrulewidth}{1pt}{0pt}
Type & Model & Model Checkpoint & Source & Reference \\
\specialrule{\lightrulewidth}{0pt}{1pt}
\multirow{6}{*}{Open-source}
& Gemma4-26B-A4B-it 
& \texttt{google/gemma-4-26B-A4B-it} 
& HuggingFace 
& \citet{google2026gemma4} \\

& \makecell[l]{Qwen3-30B-A3B-Instruct} 
& \makecell[l]{\texttt{Qwen/Qwen3-30B-A3B-Instruct-2507}} 
& HuggingFace 
& \citet{yang2025qwen3} \\

& Solar-Pro-3 
& \texttt{solar-pro3-260323} 
& Upstage 
& \citet{solarpro3_2026} \\

& \makecell[l]{Nemotron-3-Super\\-120B-A12B} 
& \makecell[l]{\texttt{nvidia/NVIDIA-Nemotron-3}\\\texttt{-Super-120B-A12B-BF16}} 
& HuggingFace 
& \citet{openai2025gpt5} \\

& \makecell[l]{Qwen3-235B-A22B-Instruct} 
& \makecell[l]{\texttt{Qwen/Qwen3-235B-A22B-Instruct-2507}}
& HuggingFace 
& \citet{yang2025qwen3} \\

& DeepSeek-V3.2 
& \texttt{deepseek-ai/DeepSeek-V3.2} 
& HuggingFace 
& \citet{liu2025deepseek} \\
\specialrule{\lightrulewidth}{1pt}{0pt}

\multirow{5}{*}{Proprietary}
& Gemini-3.1-Flash-Lite 
& \texttt{gemini-3.1-flash-lite} 
& Google API 
& \citet{google2026geminiflash} \\

& Gemini-3.1-Pro 
& \texttt{gemini-3.1-pro-preview} 
& Google API 
& \citet{google2026geminipro} \\

& GPT-5.4-mini 
& \texttt{gpt-5.4-mini-2026-03-17} 
& OpenAI API 
& \citet{openai2026gpt54} \\

& GPT-5.4 
& \texttt{gpt-5.4-2026-03-05} 
& OpenAI API 
& \citet{openai2026gpt54} \\

& o4-mini-deep-research 
& \texttt{o4-mini-deep-research-2025-06-26} 
& OpenAI API 
& \citet{openai2025deepresearch} \\
\specialrule{\heavyrulewidth}{0pt}{0pt}
\end{tabular}
\vspace{-0.25cm}
\captionof{table}{Backbone LLM configurations for \algname{}.}
\vspace{-0.3cm}
\label{tab:backbone_llm}
\end{table}

\section{Scientific Artifacts}
Our experiments use publicly accessible LLMs as party, mediator, and evaluator backbones, accessed via the OpenAI and Google APIs for proprietary models and via Hugging Face checkpoints for open-weight models under their respective terms and licenses. The exact checkpoints are listed in Appendix~\ref{app:models}.
Source scenarios are synthesized from deep-research seeds (\S\ref{sec:method:construction}) and do not incorporate text from any licensed corpus, so their use is consistent with the intended use of the underlying models and poses no licensing conflict. 

\section{Model Specifications}
\label{app:models}

Table~\ref{tab:backbone_llm} lists the LLM backbones used across all \algname{} experiments and pipeline stages: the \emph{Searcher} (o4-mini-deep-research) for seed collection, the \emph{Scenario Writer} (GPT-5.4) for recasting and condition-expansion rewrites, the \emph{Simulator} (party agents, DeepSeek-V3.2) for role-played negotiation, the benchmarked \emph{Mediators}, and the \emph{Evaluator} (DeepSeek-V3.2) for topic-localized scoring. The same pool also supplies the \emph{Fidelity Simulators} for the persona-fidelity validation (\S\ref{sec:validation:sim}): the mediator pool plus the two updated backbones GPT-5.4 and Gemini-3.1-Pro, which are used to isolate persona controllability from weak-simulator failure. Proprietary models are accessed via their official APIs and open-source models via Hugging Face checkpoints.

\section{Agentic Scenario Construction Details}
\label{app:scenario_detail}

We provide the prompt details for the three stages of the scenario construction process in \S\ref{sec:method:construction}, including seed search, scenario recasting, and the party agent used in the rejection-sampling filter.

\paragraph{Seed Search.}
Table~\ref{tab:prompt-research} is the prompt issued to the o4-mini-deep-research agent for each domain, with the domain filling the query field. The agent returns a seed report covering the conflict's timeline, stakeholders, core issues, institutional tensions, and current status. Table~\ref{tab:seed-example} shows one such seed for the Healthcare domain, drawn from a publicly documented hospital-closure dispute.

\paragraph{Scenario Recast.}
Table~\ref{tab:prompt-scenario} presents the recast prompt for the GPT-5.4 scenario writer (temperature $=0$), which converts a conflict seed into a structured scenario. The prompt enforces fictional names for every real entity, at most four topics each with a small discrete option set, and diverging party stances with at least one emotionally provocative topic. Table~\ref{tab:scenario-example} shows the scenario recast from the seed in Table~\ref{tab:seed-example}; reading the two tables side by side shows that the operator-versus-regulator pairing, the four issue clusters (emergency access continuity, offset investments at receiving hospitals, workforce protections, and accountability for premature service reductions), and the asymmetry between operator financial pressure and statutory regulatory mandate carry over from the real conflict, while all identifying names, dollar figures, and dates are replaced with fictional substitutes.

\begin{table*}[t]
\centering
\footnotesize
\renewcommand{\arraystretch}{1.2}
\setlength{\tabcolsep}{6pt}
\begin{tabularx}{\textwidth}{|l|c|X|}
\specialrule{\heavyrulewidth}{0pt}{0pt}
\multicolumn{1}{|c|}{Axis} & \multicolumn{1}{c|}{\# of Conditions} & \multicolumn{1}{c|}{Condition Pairings} \\
\specialrule{\lightrulewidth}{0pt}{0pt}
General              & 1  & Unexpanded Condition (Default) \\ 
Strategic Posture    & 3  & Competing, Avoiding, Accommodating \\
Party Composition    & 1  & Three-party \\
History Length       & 1  & Extended Length ($5\times$) \\
Emotional Reactivity & 3  & Com-Com, Com-React, React-React \\
Cultural Identity    & 6  & US-US, CN-CN, KR-KR, US-CN, US-KR, CN-KR \\
\specialrule{\lightrulewidth}{0pt}{0pt}
Total                & 15 & \\
\specialrule{\heavyrulewidth}{0pt}{0pt}
\end{tabularx}
\vspace{-0.2cm}
\caption{The 15 conditions per scenario, listed by axis. The general condition is the unexpanded baseline retained from agentic scenario curation (\S\ref{sec:method:construction}), while the remaining 14 are produced by applying one socio-cognitive axis to a fresh copy of the scenario.}
\label{tab:condition-list}
\vspace{-0.4cm}
\end{table*}

\paragraph{Preference Weighting.}
Table~\ref{tab:prompt-weight} is the prompt issued to the GPT-5.4 scenario writer to derive each party's preference weights $\mathcal{W}$ over topics and per-topic stances from its profile. Weights are positive integers summing to $100$, and the prompt forbids uniform distributions to ensure a clear priority ordering. The resulting weights and stances enter the party profile and are reused across all condition expansions, but the Parties axis is the only case that issues this prompt again, where the original parties keep their general condition weights and only the newly added party receives a fresh assignment.

\paragraph{Party Agent.}
Table~\ref{tab:prompt-party} is the prompt for the role-playing party agents (DeepSeek-V3.2, temperature $=0.6$) used in both the unmediated rejection-sampling simulation and all downstream conditions. On its turn, an agent emits a private inner thought followed by a public utterance; the inner thought is appended to the party's private history and is never visible to other parties or to the mediator. A simulation terminates as resolved once every party explicitly signals consensus, or as an impasse when a party emits an impasse signal or the turn budget is exhausted.

\section{Socio-Cognitive Probing Details}
\label{app:probing}

We detail the implementation of each component described in \S\ref{sec:method:simulation} the five condition axes, and the benchmarked mediators used for social conflict resolution benchmarking.

For each axis, we describe only the implementation mechanism and the prompt that drives it; the targeted competency and motivation are in \S\ref{sec:method:simulation}. Across the five axes plus the unexpanded baseline, every scenario is expanded into the $15$ conditions enumerated in Table~\ref{tab:condition-list}.

\paragraph{Strategies.}
We append a Thomas-Kilmann mode instruction (\emph{competing}, \emph{avoiding}, or \emph{accommodating}) to the background $\mathcal{B}$; no LLM rewrite. The three conditions differ in this instruction alone.

\paragraph{Parties.}
Using Table~\ref{tab:prompt-parties-exp}, the scenario writer revisits the real-world seed and adds one structurally distinct party with its own role, relation, and per-topic stances. Original parties and topics are carried over verbatim, so the added difficulty comes from tracking more states.

\paragraph{Histories.}
The scenario writer (Table~\ref{tab:prompt-history-exp}) prepends four dated narrative entries extracted from the seed's event sequence, with the original background appended unchanged as the final state. This expands the background to roughly five times its default length.

\paragraph{Emotion Control.}
We append a fixed reactiveness template parameterized by $r \in [0,1]$ to the party profile, contrasting volatile/escalating behavior at $r{=}1$ with calm/composed behavior at $r{=}0$; no LLM rewrite is used. For condition expansion, we use the two endpoints, composed (C, $r{=}0$) and reactive (R, $r{=}1$), forming three pairings (CC, CR, RR). Intermediate values are validated in \S\ref{sec:validation:sim}.

\paragraph{Culture.}
We anchor each party to a US, CN, or KR identity by appending to its profile a deterministic statement that summarizes the culture's 0--100 scores across the six Hofstede dimensions~\citep{hofstede2010cultures}. Pairing the three identities yields three intra-cultural and three cross-cultural conditions. All parties interact in English regardless of identity.
Each dimension ranges from 0 to 100, with higher scores indicating stronger expression of the named tendency:
\begin{itemize}[leftmargin=*, noitemsep, topsep=2pt]
    \item \emph{Power Distance}: low scores reflect flat, consensus-oriented decision-making, while high scores reflect acceptance of hierarchical authority and unequal power distribution.
    \item \emph{Individualism vs.\ Collectivism}: low scores indicate in-group loyalty and collective identity, while high scores indicate personal autonomy and self-reliance.
    \item \emph{Masculinity vs.\ Femininity}: low scores reflect cooperative, relational values, while high scores reflect competitive, performance-oriented values.
    \item \emph{Uncertainty Avoidance}: low scores indicate tolerance for ambiguity and unstructured situations, while high scores indicate a preference for clear rules and predictability.
    \item \emph{Long-term vs.\ Short-term Orientation}: low scores reflect adherence to tradition and short-term outcomes, while high scores reflect pragmatic, future-oriented planning.
    \item \emph{Indulgence vs.\ Restraint}: low scores indicate norm-compliant restraint of desires, while high scores indicate free expression of needs and enjoyment.
\end{itemize}
The US statement foregrounds individual independence and direct expression. The CN statement emphasizes relational networks, hierarchy, and long-term strategy. The KR statement shares the East Asian long-term orientation but exhibits notably higher uncertainty avoidance and a stronger preference for implicit consensus.


\section{Topic-Localized Evaluation Prompts}
\label{app:evaluator_detail}
Table~\ref{tab:prompt-eval} is the topic-localized evaluation prompt. The judge, run at temperature $0$, reads the full conversation once, identifies every turn where the topic is actively discussed or a party shifts position, and at each such turn records a $1$--$5$ agreement score together with each party's stance expressed using the topic's option labels.

\section{Validation Details}
\label{app:validation_detail}

This appendix reports the annotation protocols used to validate two components of \algname{}: persona fidelity in simulated parties and consensus alignment in the topic-localized evaluator. We summarize recruitment, task design, and quality control for each protocol.

\subsection{Simulation Fidelity Annotations}
\label{app:simulation_validation_detail}
We focus on persona fidelity because the remaining axes are either structural by construction or externally validated in prior work. Party composition and history length are direct structural perturbations, while strategic posture \citep{chen2026simulating, liu2025promediate} and cultural persona realization \citep{dey2025can} have been validated in prior LLM simulation studies. Other dimensions such as naturalness~\citep{liu2025promediate} and instruction adherence~\citep{vaccaro2025advancing} have likewise been studied and established in prior works.


\paragraph{Annotator Qualification and Compensation.}
We collect persona fidelity annotations through Amazon Mechanical Turk (MTurk), restricting participation to workers with a HIT approval rate above $90\%$, at least $500$ approved HITs, and a minimum score of $90$ on a custom English-comprehension qualification test. Annotators are compensated at \$7.50 per hour, above the U.S. federal minimum wage, and no personally identifiable information is collected.

\begin{figure*}[p]
\centering
\includegraphics[width=\textwidth]{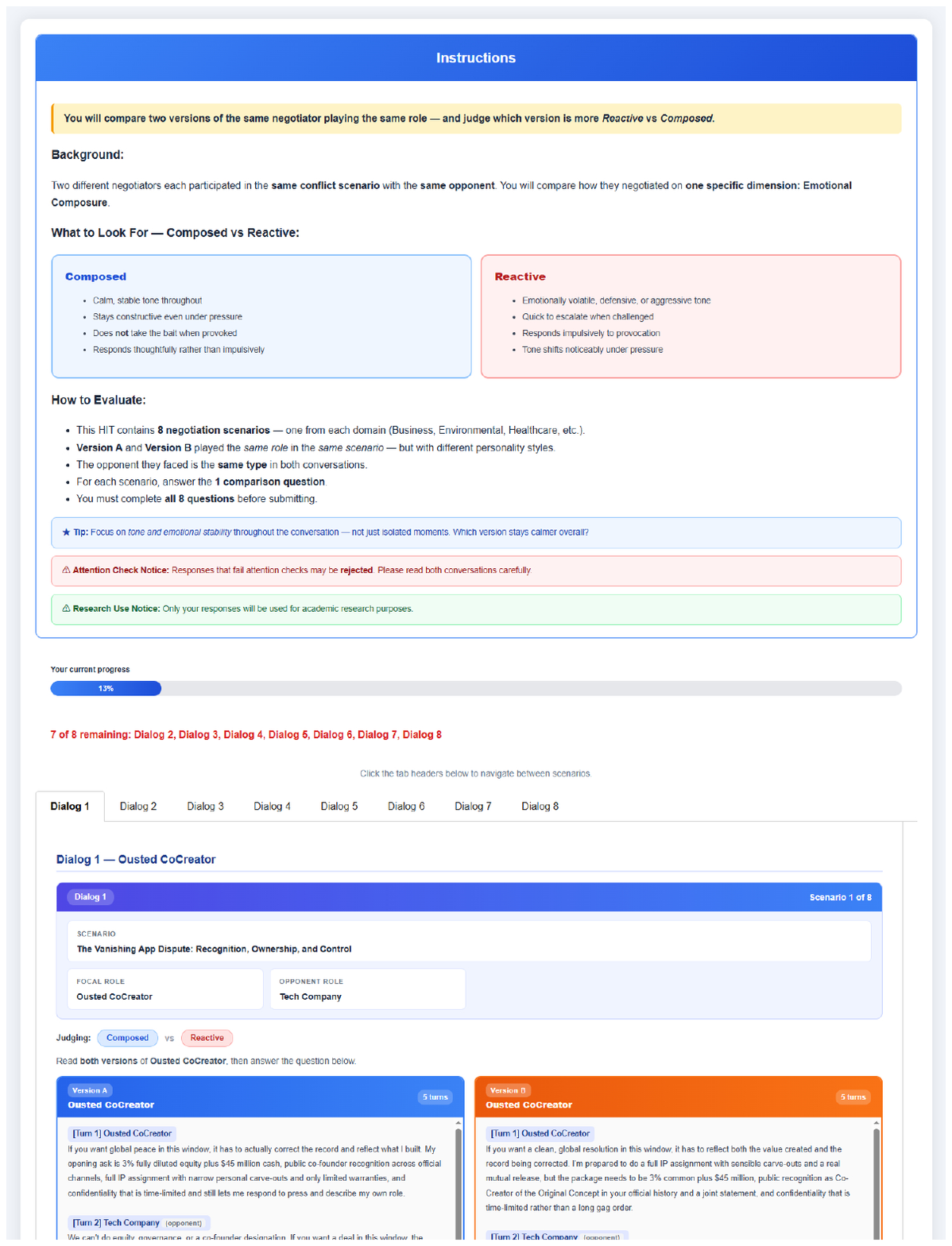}
\vspace{-0.2cm}
\caption{Example of annotation template for pairwise simulation fidelity evaluation.}
\label{fig:fidelity-template}
\vspace{-0.3cm}
\end{figure*}

\paragraph{Task Design.}
Annotators compare two conversations generated from the same scenario, party role, opponent, and topic structure, with only the target party's reactiveness level changed. They select which dialogue better reflects a \emph{reactive} rather than \emph{composed} negotiator, ignoring topic content or persuasion success unless it directly signals emotional reactivity. For each simulator backbone, we sample $160$ A/B pairs from the reactiveness grid $\{0, 0.33, 0.66, 1\}$; across seven backbones, this yields $1{,}120$ comparisons, each labeled by three annotators. Figure~\ref{fig:fidelity-template} shows the annotation template.

\paragraph{Quality Control.}
We use majority vote over the three labels for each pair and report fidelity as the fraction of pairs where the selected dialogue matches the higher assigned reactiveness level. Inter-annotator agreement is $\alpha = 0.75$, reflecting the graded nature of emotional expression, but sufficient to distinguish simulator backbones in Table~\ref{tab:persona-reactiveness}.

\subsection{Consensus Alignment Annotations}
\label{app:evaluation_validation_detail}

\paragraph{Annotator Qualification and Compensation.}
We recruit two graduate student annotators with strong English proficiency to validate consensus scoring. They are not professional negotiators. Instead, the protocol is supervised by a researcher with a graduate degree in political science and international relations, along with academic training in negotiation and diplomacy, who reviews the rubric, calibrates examples, and resolves procedural questions. As a non-expert baseline, we additionally collect three annotations for each snippet from Amazon Mechanical Turk (MTurk) workers following the same qualification protocol used in Appendix~\ref{app:simulation_validation_detail}, including a HIT approval rate above $90\%$, at least $500$ approved HITs, and a minimum score of $90$ on a custom English comprehension qualification test. Annotators are compensated for the task, no personally identifiable information is collected, and the supervised set contains $1{,}844$ snippets from $144$ conversations. The two graduate annotators reach Krippendorff's $\alpha = 0.86$.

\begin{figure*}[p]
\centering
\includegraphics[width=\textwidth]{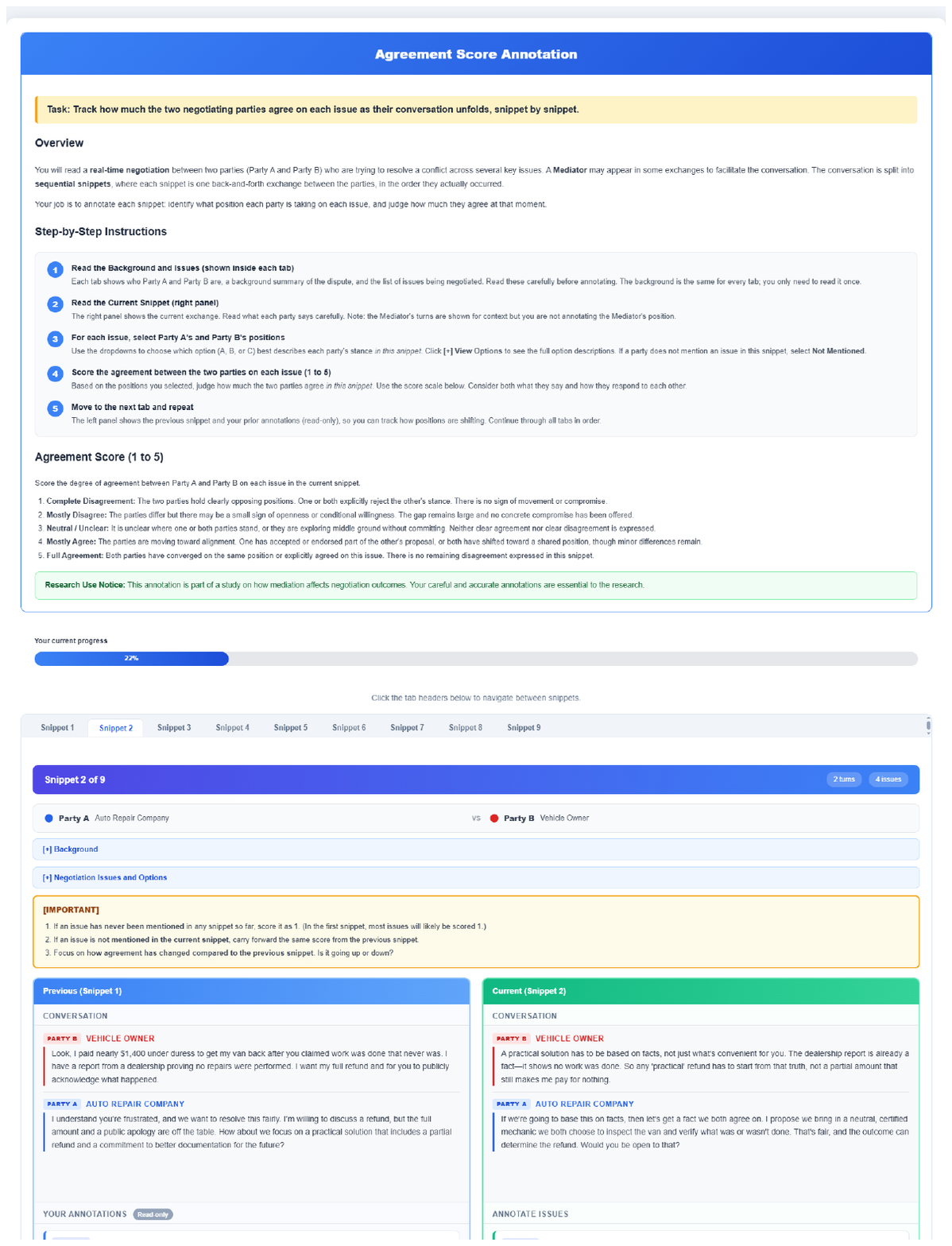}
\vspace{-0.2cm}
\caption{Example of annotation template for consensus score evaluation.}
\label{fig:alignment-template}
\vspace{-0.3cm}
\end{figure*}

\paragraph{Task Design.}
We annotate consensus at the snippet level, since agreement is interpretable only after a position receives a response. Each snippet contains one back-and-forth exchange, the background, topics, options, and the preceding snippet. For every issue, annotators record both parties' option-level positions and assign a $1$--$5$ agreement score. If an issue is not mentioned, annotators carry forward the previous score. Figure~\ref{fig:alignment-template} shows the interface.

\paragraph{Quality Control.}
The supervised graduate annotations define the reference for evaluator validation, while the non-expert annotator is retained as a baseline. Because consensus is graded, we average the two supervised annotator scores for each topic-snippet pair rather than forcing a hard adjudicated label. We then compare \algname{}, the non-expert annotator, and a per-turn LLM-judge baseline by Pearson correlation against this supervised-annotation mean at the trajectory and outcome levels.

\begin{wrapfigure}{r}{0.55\textwidth}
\vspace*{-0.5cm}
\begin{center}
\includegraphics{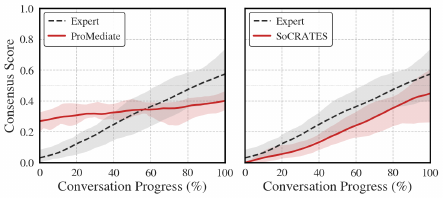}
\end{center}
\vspace*{-0.4cm}
{\small \hspace*{1.1cm} (a) \textsc{ProMediate}. \hspace*{1.1cm} (b) \algname{}.}
\vspace*{-0.2cm}
\caption{Trend comparison of consensus score trajectories for \textsc{ProMediate} and \algname{}. Bold lines show the average trajectory across dialogues, while faint lines in the background depict individual mediation trajectories, illustrating the variability across conversations.}
\label{fig:trend}
\vspace*{-0.2cm}
\end{wrapfigure}

\paragraph{Trend Comparison.} Aggregate correlations alone cannot reveal whether an evaluator tracks consensus over time. We therefore diagnose evaluators at the trajectory level, tracing how the consensus score changes over snippets and comparing it against expert annotations. A reliable evaluator should follow similar trends, showing upward progress as conflicts move toward resolution.
As shown in Figure~\ref{fig:trend}, the topic-localized evaluator (\algname{}) tracks the expert's curve closely, rising from low initial values and preserving the overall shape. In contrast, ProMediate's per-turn judge produces an unstable trajectory with large fluctuations between adjacent snippets, starting too high and ending well below the expert's final score. This instability arises because the per-turn judge scores every utterance against all issues, so inactive topics contribute uninformative scores that obscure the underlying progress. The topic-localized design evaluates only issues that are active at each moment, improving both pointwise correlation with expert judgments and the consensus dynamics underlying trajectory evaluation.

\begin{wraptable}{r!}{0.5\textwidth} 
\vspace{-0.4cm}
\centering
\renewcommand{\arraystretch}{1.1}
\footnotesize
\begin{tabular}{|L{1.7cm}|X{2cm}|X{1.9cm}|}
\specialrule{\heavyrulewidth}{1pt}{0pt}
Evaluator & Trajectory level & Outcome level \\
\specialrule{\lightrulewidth}{0pt}{0pt}
\textsc{ProMediate}        & 0.423 (0.000) & 0.394 (0.000) \\
\algname{}     & \textbf{0.785} (0.000) & \textbf{0.721} (0.000) \\
\specialrule{\heavyrulewidth}{0pt}{1pt}
\end{tabular}
\vspace{-0.2cm}
\caption{Evaluator alignment with expert judgments (Pearson $r$) using Qwen3-235B-A22B-Instruct as the backbone. Values in parentheses denote p-values.}
\label{tab:trace-validation-other}
\vspace{-0.4cm}
\end{wraptable}

\paragraph{Backbone Robustness.}
To verify that the evaluator's reliability is not tied to a specific backbone, we replace DeepSeek-V3.2 with Qwen3-235B-A22B-Instruct and re-measure alignment with expert judgments. Table~\ref{tab:trace-validation-other} shows that \algname{} preserves strong alignment under this substitution, confirming that the evaluator transfers across backbones rather than relying on a single model's behavior.

\section{Mediator Prompts}
\label{app:mediator_detail}

At every party turn, the mediator, run at temperature $0.6$, first executes the \emph{when-to-intervene} decision (Table~\ref{tab:prompt-mwhen}). If the decision is true, the \emph{how-to-intervene} generation step (Table~\ref{tab:prompt-mhow}) emits a single utterance, which is inserted before the next party speaks.

\section{Additional Analysis}
\label{app:intervention_analysis}

\subsection{Intervention Analysis}
\label{app:intervention_analysis}

\begin{wraptable}{r!}{0.5\textwidth} 
\vspace{-0.4cm}
\centering
\renewcommand{\arraystretch}{1.1}
\footnotesize
\setlength{\tabcolsep}{7pt}
\begin{tabular}{|l|l|c|c|}
\specialrule{\heavyrulewidth}{0pt}{0pt}
Type & Mediator & IF (\%) & FI (\%) \\
\specialrule{\lightrulewidth}{0pt}{0pt}
\multirow{2}{*}{Prop.}
 & Gemini-3.1-FL & 22.6 & 32.3 \\
 & GPT-5.4-mini & 22.6 & 31.0 \\
\specialrule{\lightrulewidth}{0pt}{0pt}
\multirow{6}{*}{\makecell[l]{Open\\Source}}
 & DeepSeek-V3.2 & 16.1 & 42.8 \\
 & Qwen3-235B & 20.8 & 39.5 \\
 & Nemotron-3-120B & 14.6 & 45.6 \\
 & Solar-Pro-3 & 32.3 & 26.9 \\
 & Gemma-4-26B & 16.4 & 37.3 \\
 & Qwen3-30B & 31.1 & 25.3 \\
\specialrule{\heavyrulewidth}{0pt}{1pt}
\end{tabular}
\vspace{-0.1cm}
\caption{Intervention behaviors of eight mediators. IF: Intervention Frequency, FI: First Intervention.}
\vspace{-0.3cm}
\label{tab:intervention-freq}
\end{wraptable}

To diagnose the gap between intervention timeliness and consensus gain, we measure two aspects of mediator behavior: Intervention Frequency (the fraction of party turns on which the mediator chooses to speak) and First Intervention (the relative turn position of the mediator's first utterance). Table~\ref{tab:intervention-freq} reports both, aggregated across all conditions. Solar-Pro-3 and Qwen3-30B intervene roughly twice as often as the top mediators and begin speaking much earlier in the conversation. This over-eager speaking inflates intervention timeliness without translating into intervention effectiveness or consensus gain, suggesting that early and frequent intervention does not substitute for substantive contribution to social conflict resolution.

\subsection{Benchmark Stability Analysis}
\label{app:stability}

We test the benchmark along three axes left fixed in the main results: the evaluator backbone, the party-agent simulator backbone, and run-to-run stochasticity. 

\paragraph{Evaluator Backbone Robustness.}
\label{app:other_evaluator}

\begin{wraptable}{r!}{0.55\textwidth} 
\vspace{-0.4cm}
\centering
\renewcommand{\arraystretch}{1.15}
\footnotesize
\setlength{\tabcolsep}{5pt}
\begin{tabular}{|l|c|c|c|c|}
\specialrule{\heavyrulewidth}{1pt}{0pt}
Metric & DS & Qw & $\Delta$ & $\rho$ \\
\specialrule{\lightrulewidth}{0pt}{0pt}
Intervention Timeliness    & 79.2 & 77.2 & $-2.0$ & 0.406 \\
Intervention Effectiveness & 21.3 & 25.2 & $+3.9$ & 0.862 \\
Consensus Gain             & 25.9 & 26.5 & $+0.6$ & 0.786 \\
\specialrule{\heavyrulewidth}{0pt}{1pt}
\end{tabular}
\vspace{-0.1cm}
\caption{Metric values averaged across mediators under two evaluator backbones (DS = DeepSeek-V3.2, Qw = Qwen3-235B-A22B-Instruct), where $\Delta$ reports Qw $-$ DS and $\rho$ denotes the Spearman correlation computed per metric over the per-scenario pairs.}
\label{tab:stab-evaluator}
\vspace{-0.2cm}
\end{wraptable}

To check whether the mediator ranking depends on the evaluator backbone, we swap DeepSeek-V3.2 with Qwen3-235B-A22B-Instruct and re-evaluate the mediation trajectories from~\S\ref{sec:experiments} again, holding the disputant simulator fixed. Table~\ref{tab:stab-evaluator} reports the average across mediators under both evaluators. The two evaluators yield close averages, differing by only $-2.0$, $+3.9$, and $+0.6$ points across the three metrics. The mediator rankings also agree well on intervention effectiveness (Spearman $\rho = 0.862$) and consensus gain ($\rho = 0.786$). Intervention Timeliness shows weaker agreement ($\rho = 0.406$), because it depends on which turn the trajectory is sampled at and is more sensitive to the evaluator's choice of relevant turns. We note that Qwen3-235B-A22B-Instruct itself is a weaker evaluator than DeepSeek-V3.2 in our validation (Table~\ref{tab:trace-validation-other}), which likely accounts for part of this gap. Even so, the relative ordering of mediators is preserved under the alternative evaluator.

\paragraph{Simulator Backbone Robustness.}
\label{app:other_simulator}

To check whether mediator adaptation across social-cognitive axes depends on the disputant simulator, we replace DeepSeek-V3.2 party agents with Qwen3-235B-A22B-Instruct. Due to the cost of simulating disputants and our limited budget, this ablation covers three mediators (Qwen3-235B, DeepSeek-V3.2, Qwen3-30B), while still spanning all $8$ situations ($600$ scenarios) used in the main experiments. 
Since this ablation includes only three representative mediators, it is not intended to support a full mediator ranking. Instead, our goal is to test whether the gaps across axes identified in \S\ref{sec:experiments:socio} persist under an alternative simulator.

\begin{wrapfigure}{r!}{0.55\textwidth} 
\begin{center}
\includegraphics{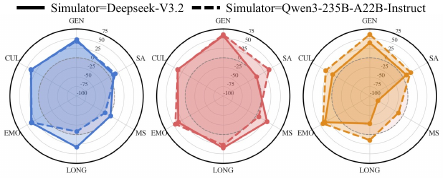}
\end{center}
\vspace*{-0.4cm}
{\small \hspace*{0cm} (a) DeepSeek-V3.2. \hspace*{0.1cm} (b) Qwen3-235B. \hspace*{0.2cm} (c) Qwen3-30B.}
\vspace*{-0.2cm}
\caption{Mediator adaptation of three mediators under two disputant simulators (DeepSeek-V3.2, solid line; Qwen3-235B-A22B-Instruct, dashed line).}
\label{fig:stab-simulator}
\vspace*{-0.3cm}
\end{wrapfigure}

As shown in Figure~\ref{fig:stab-simulator}, absolute consensus gain values shift after the simulator swap, but both the shape of each mediator's adaptation profile and the distinctions among mediators are largely preserved. Under both simulators, the general condition remains the strongest setting, and performance drops when moving to the perturbed axes. The relative pattern across axes is preserved under the alternative simulator. Cultural adaptation shows a milder decline, whereas tracking multiple party states and using long histories show larger degradations. The three mediators also retain their characteristic profiles under the alternative simulator, rather than collapsing to a common pattern. This consistency suggests that the social and cognitive gaps measured by \algname{} reflect adaptation limits of each mediator rather than artifacts of a particular disputant simulator.

\begin{table*}[t!]
\centering
\renewcommand{\arraystretch}{1.1}
\footnotesize
\setlength{\tabcolsep}{12pt}
\begin{tabular}{|l|l|c|c|c|}
\specialrule{\heavyrulewidth}{1pt}{0pt}
Type & Mediator & \makecell{Intervention\\Timeliness} & \makecell{Intervention\\Effectiveness} & Consensus Gain \\
\specialrule{\lightrulewidth}{0pt}{0pt}
\multirow{2}{*}{\makecell[l]{Proprietary}}
 & Gemini-3.1-FL      & $76.3 \pm 0.5$  & $21.2 \pm 1.4$ & $46.6 \pm 2.5$  \\
 & GPT-5.4-mini       & $78.5 \pm 2.4$  & $20.8 \pm 0.3$ & $48.4 \pm 1.6$  \\
\specialrule{\lightrulewidth}{0pt}{0pt}
\multirow{6}{*}{\makecell[l]{Open-source}}
 & DeepSeek-V3.2      & $72.4 \pm 4.4$  & $21.4 \pm 0.4$ & $50.0 \pm 2.9$  \\
 & Qwen3-235B         & $73.0 \pm 2.1$  & $24.1 \pm 0.4$ & $55.8 \pm 1.7$  \\
 & Nemotron-3-120B    & $71.4 \pm 2.0$  & $13.4 \pm 3.5$ & $35.4 \pm 6.8$  \\
 & Solar-Pro-3        & $79.6 \pm 1.0$  & $11.7 \pm 2.7$ & $24.5 \pm 5.6$  \\
 & Gemma-4-26B        & $71.9 \pm 0.5$  & $17.1 \pm 1.8$ & $44.4 \pm 2.7$  \\
 & Qwen3-30B          & $78.3 \pm 0.6$  & $15.4 \pm 1.1$ & $40.3 \pm 1.2$  \\
\specialrule{\heavyrulewidth}{0pt}{1pt}
\end{tabular}
\vspace{-0.1cm}
\caption{Intervention timeliness, Intervention effectiveness, and consensus gain across three independent runs on the general scenario, reported as median $\pm$ half-range.}
\vspace{-0.3cm}
\label{tab:stab-variance}
\end{table*}

\paragraph{Multi-run Robustness.}
\label{app:run_variance}
To estimate variance across multiple runs, we repeat the mediator phase two additional times for all 8 mediators, yielding three independent runs whose variance reflects both the mediator and the disputant simulator. Due to the cost of simulating the disputants and our limited budget, this ablation is limited to the general conditions. Table~\ref{tab:stab-variance} reports the median and half-range of each mediator across the three runs. On the consensus gain ranking, the three runs yield a Kendall's $W$ of 0.929, indicating strong agreement on the relative ordering of the eight mediators. Six of the eight mediators stay within a half-range of $\pm 3$ points across runs, and the remaining variance is concentrated in the lowest-ranked models. Overall, the mediator ranking is robust to repeated runs, confirming that our main findings reflect genuine mediator differences rather than stochastic noise.

\clearpage

\begin{table*}[p]
\centering
\begin{promptbox}{Seed Search Prompt}
\begin{userbox}
\scriptsize
Research real-world conflicts and disputes related to the following query, and provide a detailed conflict analysis.

\smallskip
\textbf{Query.} \textcolor{blue}{\{Query\}}

\smallskip
Search the web for notable, well-documented real-world conflicts matching this query. Pick \emph{one} specific, real conflict that would make a rich negotiation simulation, with multiple stakeholders and multiple issues.

\smallskip
\textbf{Provide.}
\begin{enumerate}[leftmargin=*, noitemsep, topsep=2pt]
\item \textbf{The specific conflict chosen}, named clearly.
\item \textbf{Timeline of key events}, the major milestones and turning points.
\item \textbf{Key stakeholders}, the main parties, roles, and interests ($3$--$5$ parties).
\item \textbf{Core issues of disagreement}, the main points of contention.
\item \textbf{Institutional tensions}, organizational, legal, or structural conflicts.
\item \textbf{Current status}, where things stand now.
\end{enumerate}

\smallskip
Be thorough and factual. Focus on structural aspects suitable for a negotiation simulation.
\end{userbox}
\end{promptbox}
\vspace{-0.25cm}
\caption{Prompt for seed search.}
\vspace{-0.3cm}
\label{tab:prompt-research}
\end{table*}

\begin{table*}[p]
\centering
\begin{promptbox}{Scenario Recast Prompt}
\begin{systembox}
\scriptsize
You are an expert negotiation simulation designer.

\smallskip
You are given RESEARCH on a real-world conflict case. Use it for structural inspiration (power asymmetries, interdependencies, topic linkages, leverage), but use fictional names for all entities, people, places, laws, and agencies.

\smallskip
\textbf{Task.} Design a negotiation simulation scenario as a JSON object, drawing on the real-world conflict case provided by the user.

\smallskip
\textbf{Party design.}
\begin{itemize}[leftmargin=*, noitemsep, topsep=2pt]
\item Design 2 parties inspired by the real conflict's stakeholder dynamics.
\item \texttt{name} captures the essence of the party (e.g.\ ``Workers Federation'', ``City Government'') without real names or acronyms.
\item \texttt{role} specifies (1) the primary objective and target outcome, (2) internal pressures or constraints, and (3) the BATNA if talks fail.
\item \texttt{relation} states who the party aligns or clashes with and on which topics.
\item \texttt{preferences} gives, for each topic, a 1--2 sentence stance. Different parties should hold different stances to create negotiation tension.
\end{itemize}

\smallskip
\textbf{Topic design.}
\begin{itemize}[leftmargin=*, noitemsep, topsep=2pt]
\item Use only the topics needed to capture the core conflict (up to 4).
\item Provide 2 options for binary positions; more only for inherently multi-level decisions.
\item Options must be concrete and substantively different; use specific numbers where natural.
\item At least one topic must be emotionally provocative and fit naturally into the scenario.
\end{itemize}
\end{systembox}

\begin{userbox}
\scriptsize
\textbf{Research Result.} \textcolor{blue}{\{Seed Scenario\}}

\smallskip
Return only valid JSON with the schema below, with no markdown fencing, preamble, or commentary.
\begin{verbatim}
{
  "title": "string",
  "background": "string (context, history, stakes;
     generic names; specific amounts/timelines)",
  "topics": [
    {"id": "string (short code, e.g. GOV)",
     "name": "string (full descriptive name)",
     "description": "string (what is at stake)",
     "options": [
       {"label": "string (letter label)",
        "description": "string (concrete action)"}]}
  ],
  "parties": [
    {"id": "string (short code, e.g. WORK)",
     "name": "string (full name)",
     "role": "string (objective, constraints, BATNA)",
     "relation": "string (ally/rival, with whom)",
     "preferences": {
       "TOPIC_ID": "string (1-2 sentence stance)"}}
  ]
}
\end{verbatim}
\end{userbox}
\end{promptbox}
\vspace{-0.25cm}
\caption{Prompt for scenario recast.}
\vspace{-0.3cm}
\label{tab:prompt-scenario}
\end{table*}
\clearpage

\begin{table*}[p]
\centering
\tiny
\renewcommand{\arraystretch}{1.15}
\begin{tabularx}{\textwidth}{@{}l X@{}}
\toprule
\textbf{Conflict} & \textcolor{blue}{Mount Sinai Beth Israel} hospital downsizing and closure, New York City (\textcolor{blue}{2016}--present). \textcolor{blue}{Mount Sinai Beth Israel (MSBI)} is a roughly 700-bed acute-care hospital in Manhattan's East Village, operated by \textcolor{blue}{Mount Sinai Health System}, a \textcolor{blue}{private nonprofit} serving the Lower East Side, East Village, and Chinatown neighborhoods. \\
\midrule
\textbf{Timeline of key events} &
2013: Mount Sinai Health System forms through merger with Continuum Health Partners, absorbing Beth Israel Medical Center.
\newline
\textcolor{blue}{2016}-05: Mount Sinai announces a \textcolor{blue}{``Downtown Transformation''} plan, an approximately \textcolor{blue}{\$500 million} investment to replace the large inpatient hospital with a smaller acute-care facility ($\sim$70 beds) at the Phillips Ambulatory Care site plus an expanded \textcolor{blue}{hub-and-spoke outpatient network} across downtown Manhattan.
\newline
2016--2022: Plan repeatedly delayed under regulatory and community pressure; Community Board 3, local elected officials, and patient advocacy groups raise objections about loss of inpatient psychiatry, addiction services, and 24/7 emergency care access.
\newline
\textcolor{blue}{2023-09}: Mount Sinai files an updated \textcolor{blue}{closure plan} with the New York State Department of Health to fully shut Beth Israel; community groups respond with Article 78 litigation and public protests.
\newline
2024: NY DOH holds public hearings on the revised closure application; preliminary conditions require Mount Sinai to maintain certain emergency and behavioral health services during transition. \\
\midrule
\textbf{Key stakeholders} &
\textcolor{blue}{\textbf{Mount Sinai Health System (MSHS)}}: \textcolor{blue}{Private nonprofit operator}; reports persistent operating losses at Beth Israel cited at \textcolor{blue}{roughly \$150M per year}; primary objective is to complete the wind-down to stem ongoing losses; BATNA is unilateral filing subject to NY DOH closure-approval authority.
\newline
\textcolor{blue}{\textbf{New York State Department of Health (NY DOH)}}: \textcolor{blue}{Statewide regulator with statutory Certificate of Need authority over hospital closures}; concerned with continuity of essential services, especially behavioral health and addiction treatment for downtown Manhattan.
\newline
\textbf{Coalition to Save Beth Israel and Community Board 3}: Local advocacy coalition representing patients and residents; demand community benefit agreements and public accountability; pursue Article 78 litigation.
\newline
\textbf{1199SEIU and New York State Nurses Association (NYSNA)}: Unions representing several thousand affected clinical and support staff; demand placement guarantees within Mount Sinai's other facilities, retraining funds, and severance protections.
\newline
\textbf{Surrounding receiving systems (Bellevue, NYU Langone, NewYork-Presbyterian Lower Manhattan)}: Hospitals expected to absorb deflected emergency, psychiatric, and inpatient volume; seek capital support from Mount Sinai to expand capacity. \\
\midrule
\textbf{Core issues of disagreement} &
\textcolor{blue}{\textbf{Continuity of emergency and behavioral health access}}: Mount Sinai favors rapid downsizing; NY DOH and community advocates demand a staged transition with continuing 24/7 emergency department capacity and crisis stabilization, citing Beth Israel's role as a primary psychiatric receiving site for downtown Manhattan.
\newline
\textcolor{blue}{\textbf{Offset investments at receiving hospitals}}: NY DOH conditions closure approval on Mount Sinai funding emergency-capacity expansion and EMS upgrades at Bellevue and other surrounding systems; Mount Sinai contests the scope and duration of these obligations.
\newline
\textcolor{blue}{\textbf{Workforce transition protections}}: Unions demand redeployment within the Mount Sinai system at equal pay, multi-year retraining funds, and enhanced severance; Mount Sinai offers severance close to statutory minima.
\newline
\textcolor{blue}{\textbf{Accountability for premature service reductions}}: NY DOH and community groups have documented quiet service reductions (inpatient psychiatry, addiction treatment beds, obstetrics) preceding formal approval; demands include public acknowledgment, financial penalties, and community benefit reinvestment from the resulting downtown real estate. \\
\midrule
\textbf{Institutional tensions} &
NY DOH's Certificate of Need authority over closures conflicts with the operator's fiduciary autonomy to manage finances. Mount Sinai's 501(c)(3) nonprofit mission obligations and Medicaid commitments conflict with mounting operating losses. The medical-school affiliation (Icahn School of Medicine at Mount Sinai) complicates residency redistribution timelines. Local elected officials and Community Board 3 carry political weight but hold no formal closure-approval authority. Receiving hospitals operate under separate Certificate of Need processes that lag the closure timeline. \\
\midrule
\textbf{Current status} &
The revised closure plan remains under NY DOH review. Beth Israel continues to operate at reduced inpatient capacity. The replacement ambulatory and acute-care site is partially operational. Active negotiation focuses on the magnitude and duration of offset investments at receiving hospitals, the scope of workforce protections, and accountability measures for documented premature reductions. \\
\bottomrule
\end{tabularx}
\caption{Example deep-research seed for the Healthcare domain, returned by the Searcher for a hospital-closure query. This seed is the input the Scenario Writer recasts into the structured scenario in Table~\ref{tab:scenario-example}. \textcolor{blue}{Blue} marks elements that carry over to the recast under fictional names.}
\label{tab:seed-example}
\end{table*}

\begin{table*}[t]
\centering
\tiny
\renewcommand{\arraystretch}{1.15}
\begin{tabularx}{\textwidth}{@{}l X@{}}
\toprule
\textbf{Title} & Downtown General Wind-Down: Regulator--Provider Bargaining Over Access, Capacity, and Accountability \\
\midrule
\textbf{Background} & A \textcolor{blue}{private nonprofit} system, \textcolor{blue}{Regional Health Network (RHN)}, operates \textcolor{blue}{Downtown General Hospital (DGH)} in the River District of Eastborough City. RHN reports sustained operating losses of \textcolor{blue}{\$120--\$150 million per year} at DGH since 2019, with average inpatient occupancy falling below 45\% and increasing reliance on short-stay and outpatient care. In \textcolor{blue}{2016}, RHN announced a \textcolor{blue}{\$520 million} \textcolor{blue}{``Downtown Transformation''} to pivot from a large inpatient footprint to a smaller \textcolor{blue}{hub-and-spoke outpatient network}; community groups pushed back, arguing the plan would hollow out emergency and behavioral health services. In \textcolor{blue}{September 2023}, RHN publicly signaled intent to close DGH and filed a formal \textcolor{blue}{closure plan} with\dots \\
\midrule
\multicolumn{2}{@{}l}{\textbf{Parties}} \\
\quad \textcolor{blue}{\texttt{RHN} (Regional Health Network)} & You are the \textcolor{blue}{private nonprofit operator} of Downtown General Hospital. Your primary objective is to complete the wind-down while shifting care to a lower-cost outpatient model, protecting system finances, and preserving your brand in the city. Constraints: sustained \textcolor{blue}{\$120--\$150 million/year losses} at DGH, bond covenant\dots \\
\quad \textcolor{blue}{\texttt{SHOA} (State Health Oversight Agency)} & You are the \textcolor{blue}{statewide regulator charged with approving the closure} and safeguarding access to essential services. Your primary objective is to secure enforceable mitigations that maintain timely emergency and behavioral health access and protect the workforce during transition. Constraints: statutory due-process\dots \\
\midrule
\multicolumn{2}{@{}l}{\textbf{Topics}} \\
\quad \textcolor{blue}{\texttt{ACC} (Continuity of Emergency Access and Urgent Care Model)} & (A)~Operate a 24/7 urgent and primary care hub within 0.3 miles of the former emergency department for 5 years, co-locating a behavioral health crisis stabilization unit (6 chairs),\dots \newline (B)~Operate a 24/7 urgent care for 3 years with 6 observation bays and on-call behavioral health; performance targets include 40-minute average transfer time; \$9 million/year RHN\dots \newline (C)~Operate a 16-hour daily urgent care for 2 years, no observation bays; after-hours coverage by tele-triage and ambulance diversion to nearby hospitals; \$5 million/year RHN\dots \newline (D)~Maintain a micro-hospital at Seaport Pavilion with a licensed 30-bed unit and full-service emergency department for 3 years during transition; RHN funds \$45 million in capital\dots \\
\addlinespace[2pt]
\quad \textcolor{blue}{\texttt{INV} (Offset Investments to Expand Regional Emergency Capacity)} & (A)~RHN transfers \$70 million in escrowed capital to CityCare Medical Center to add 30 emergency department treatment positions, 8 fast-track bays, and retrofit 2 negative-pressure\dots \newline (B)~RHN funds \$40 million to CCMC for a 20-position expansion plus \$10 million to upgrade emergency medical services dispatch, radios, and 4 new ambulances; 5-year service covenant\dots \newline (C)~RHN establishes a \$25 million Community Access Fund for care coordination, behavioral health integration, and transport vouchers; no direct emergency department capital expansion. \newline (D)~No new investments; rely on existing regional capacity and RHN's outpatient network to handle deflected demand. \\
\addlinespace[2pt]
\quad \textcolor{blue}{\texttt{WORK} (Workforce Transition Protections)} & (A)~Guarantee placement within 25 miles for at least 85\% of affected full-time equivalent roles at equal or higher base pay for 24 months; \$12 million retraining fund; \$20,000\dots \newline (B)~Guarantee first-right-of-hire at RHN sites with pay protection for 12 months; severance of 3 weeks per year of service capped at 52 weeks; \$6 million retraining fund; private\dots \newline (C)~Statutory minimum severance only; \$2 million training vouchers; no redeployment guarantees; internal reporting. \\
\addlinespace[2pt]
\quad \textcolor{blue}{\texttt{ACCNT} (Accountability and Public Narrative About Premature Service Reductions)} & (A)~RHN's chief executive issues a public apology acknowledging anxiety and access risks caused by premature reductions; fund a \$5 million Community Stabilization Grant program\dots \newline (B)~Issue a joint statement of shared responsibility with SHOA; appoint an independent reviewer with public quarterly reports for 2 years; provide \$1 million in transport vouchers\dots \newline (C)~Adopt a no-fault, forward-looking compliance plan with internal reporting and SHOA access to records; no apology, no public audit, and no community grants. \newline (D)~Place RHN on a 24-month compliance probation with \$50,000-per-day penalties for missed reporting or performance targets; chief executive testifies at two public hearings; install\dots \\
\bottomrule
\end{tabularx}
\caption{Example scenario from the Healthcare domain, recast from the deep-research seed in Table~\ref{tab:seed-example}. \textcolor{blue}{Blue} marks elements paired with the corresponding blue elements in the seed.}
\label{tab:scenario-example}
\end{table*}
\clearpage

\begin{table*}[p]
\centering
\begin{promptbox}{Preference Weighting}
\begin{systembox}
\scriptsize
You are an expert negotiation simulation designer.

\smallskip
You are given a SCENARIO with its title, background, issues, and parties; each party may carry persona attributes (cultural background, emotion control).

\smallskip
\textbf{Task.} For each party, generate (1) preference \texttt{weights} as positive integers summing to exactly $100$ across all issues, and (2) a 1--2 sentence \texttt{stance} per issue stating what the party wants and why.

\smallskip
\textbf{Rules.}
\begin{itemize}[leftmargin=*, noitemsep, topsep=2pt]
\item Weights must be positive integers summing to exactly $100$; allow uneven distributions and do not spread weight evenly across issues.
\item A party's structural position, objectives, constraints, and BATNA are the main drivers of both weights and stances.
\end{itemize}
\end{systembox}

\begin{userbox}
\scriptsize
\textbf{Scenario.} \textcolor{blue}{\{Background\}}

\smallskip
\textbf{Topics.} \textcolor{blue}{\{Topics\}}

\smallskip
\textbf{Parties.} \textcolor{blue}{\{Parties\}}

\smallskip
Return only valid JSON with the schema below, with no markdown fencing, preamble, or commentary.
\begin{verbatim}
{
  "PARTY_ID": {
    "weights": {"ISSUE1": int, "ISSUE2": int, ...},
    "stances": {"ISSUE1": "wants X because Y...",
                "ISSUE2": "prefers Z due to..."}}
}
\end{verbatim}
\end{userbox}
\end{promptbox}
\vspace{-0.25cm}
\caption{Prompt for per-party preference weighting.}
\vspace{-0.3cm}
\label{tab:prompt-weight}
\end{table*}

\begin{table*}[p]
\centering
\begin{promptbox}{Conflict Simulation: party agent (inner thought and utterance)}

\begin{systembox}
\scriptsize
You are \textcolor{blue}{\{Name\}} (\textcolor{blue}{\{Role\}}), negotiating with \textcolor{blue}{\{Opponent\}}. Your goal is to reach the best possible outcome for yourself.

\smallskip
\textbf{Task.} Generate one internal thought ($\le$2 sentences) reflecting what is on your mind right now, then write your next spoken utterance based on that thought.
\begin{itemize}[leftmargin=*, noitemsep, topsep=2pt]
\item The thought is your private reasoning; stay grounded in your positions and prior thoughts.
\item The utterance is your actual turn: concise speech based on your current thought and the conversation so far.
\item Think and speak as a real person: fully embody your persona (personality, cultural background) and conflict mode.
\item Negotiate across multiple topics; consider the overall deal, not one topic at a time.
\item Pursue your interests without telegraphing your strategy or priorities directly.
\item Use plain language, with no speaker labels, headers, option labels, or bullet points.
\item Closing protocol: end with \texttt{[IMPASSE]} if no agreement is possible; write \texttt{[FINAL AGREEMENT]} only after the other party has explicitly accepted your exact terms.
\end{itemize}

\end{systembox}

\begin{userbox}
\scriptsize
\textbf{Background.} \textcolor{blue}{\{Background\}} 
\textcolor{blue}{\{Conflict Mode Instruction\}}

\smallskip
\textbf{Previous thoughts.} \textcolor{blue}{\{Previous Thoughts\}}

\smallskip
\textbf{Persona.} 
\textcolor{blue}{\{Party Profile\}} 
\textcolor{blue}{\{Persona Instruction\}}

\smallskip
\textbf{Your interests (100 pts total).} \textcolor{blue}{\{Preferences\}} Pursue high-weight topics more and flex on low-weight ones; express interests through questions and framing, not by stating priorities directly.

\smallskip
\textbf{Negotiable topics.} \textcolor{blue}{\{Topics\}}

\smallskip
\textbf{Conversation.} \textcolor{blue}{\{Conversation Log\}}

\smallskip
Respond only with JSON:
\begin{verbatim}
{ "thought": "your internal thought here",
  "utterance": "your spoken utterance here" }
\end{verbatim}
\end{userbox}
\end{promptbox}
\vspace{-0.25cm}
\caption{Prompt for persona-conditioned party simulation.}
\vspace{-0.3cm}
\label{tab:prompt-party}
\end{table*}
\clearpage

\begin{table*}[p]
\centering
\begin{promptbox}{Parties Expansion}
\begin{userbox}
\scriptsize
You are an expert negotiation simulation designer.

\smallskip
You are given (1) a \textbf{base scenario} with 2 parties and a set of issues, to preserve as the foundation, and (2) the \textbf{research} on the real-world conflict that inspired it, used to identify additional stakeholders.

\smallskip
\textbf{Task.} Expand the scenario from 2 parties to exactly \textcolor{blue}{\{Targeted Party\}} parties, preserving the original structure.

\smallskip
\textbf{Party expansion.}
\begin{itemize}[leftmargin=*, noitemsep, topsep=2pt]
\item Keep the original 2 parties exactly as they are; only their \texttt{relation} or \texttt{preferences} may be updated to reference newly added parties.
\item Add exactly \textcolor{blue}{\{Targeted Party\}} new part(ies) inspired by real stakeholders in the research, with structurally distinct roles, not subdivisions of existing parties.
\item \texttt{name} captures the party's essence without real names or acronyms; \texttt{role} specifies its primary objective and target outcome, internal pressures or constraints, and BATNA; \texttt{relation} states whom it aligns or clashes with and on which issues.
\item \texttt{preferences} gives, for each issue, a 1--2 sentence stance following from the party's role and constraints; different parties should hold different stances to create negotiation tension.
\item Expand, do not replace, the \texttt{background} to introduce the new parties.
\end{itemize}

\smallskip
\textbf{Issues.} Keep all original issues exactly as they are (id, name, description, options); do not modify existing issues or add new ones.

\smallskip
\textbf{Base scenario.} \textcolor{blue}{\{Recast Scenario\}}

\smallskip
\textbf{Research.} \textcolor{blue}{\{Seed Scenario\}}

\smallskip
Return only valid JSON with the same schema as the base scenario (\texttt{title}, \texttt{background}, \texttt{issues}, \texttt{parties}), with no markdown fencing, preamble, or commentary.
\end{userbox}
\end{promptbox}
\vspace{-0.25cm}
\caption{Prompt for party-axis expansion.}
\vspace{-0.3cm}
\label{tab:prompt-parties-exp}
\end{table*}

\begin{table*}[p]
\centering
\begin{promptbox}{Histories Expansion}
\begin{userbox}
\scriptsize
You are an expert negotiation simulation designer.

\smallskip
You are given (1) a \textbf{base scenario} with 2 parties and a set of issues, (2) the \textbf{research} on the real-world conflict that inspired it, and (3) the \textbf{original background}, the current state of affairs the parties negotiate from, which must not be rewritten or replaced.

\smallskip
\textbf{Task.} Expand a series of historical entries that chronologically lead up to the original background. These entries are the pre-history, the events, decisions, and escalations that explain how and why the parties arrived at the current impasse, so that a reader finishing the last entry feels the original background is its natural, inevitable result.

\smallskip
\textbf{Rules.}
\begin{itemize}[leftmargin=*, noitemsep, topsep=2pt]
\item Produce exactly \textcolor{blue}{\{N Events\}} history entries; the final entry must hand off directly into the original background.
\item Use fictional, essence-capturing names for all parties; include specific numbers, dollar amounts, percentages, headcounts, dates, and durations.
\item Do not contradict, reproduce, or summarize the original background; stop before it.
\end{itemize}

\smallskip
\textbf{Structure.} For each key event extracted from the research, write a detailed entry headed \texttt{\#\#\# History [N]: [Title] ([Date])} followed by a rich narrative of what happened, why it mattered, and how it shifted the dynamics between parties. Entries follow chronological order: early entries cover origins and the status quo, middle entries cover triggering events and escalation, and late entries cover the final breakdown that directly sets up the original background.

\smallskip
\textbf{Base scenario.} \textcolor{blue}{\{Recast Scenario\}} \quad \textbf{Research.} \textcolor{blue}{\{Seed Scenario\}} \quad \textbf{Original background.} \textcolor{blue}{\{Original Background\}}

\smallskip
Return only the history entries with their \texttt{\#\#\# History [N]} headers, with no JSON wrapping, markdown fencing, preamble, or commentary.
\end{userbox}
\end{promptbox}
\vspace{-0.25cm}
\caption{Prompt for history-axis expansion.}
\vspace{-0.3cm}
\label{tab:prompt-history-exp}
\end{table*}
\clearpage

\begin{table*}[p]
\centering
\begin{promptbox}{Mediator: when-to-intervene decision}
\begin{systembox}
\scriptsize
You are the Mediator in a multi-party negotiation. Your role is to decide whether to send a message to the participants at this moment, based on the conversation history and scenario context.

\smallskip
\textbf{Task.}
\begin{itemize}[leftmargin=*, noitemsep, topsep=2pt]
\item Decide whether to send a message to the participants at this moment.
\item Stay sensitive to the social dynamics of the conversation and the participants' sentiments.
\item Be proactive in offering help, but avoid interrupting the flow following the criteria below.
\item Use plain language, with no turn labels, headers, option labels, or bullet points.
\end{itemize}

\smallskip
\textbf{Engagement criteria.} Do \emph{not} engage if the conversation is flowing well, the participants are having a personal exchange, or you are unsure whether engaging is appropriate. \emph{Engage} if the conversation has stalled or drifted from the goal, there is confusion or misalignment you can resolve, emotional tension is escalating, or a participant asked a question you can help with.
\end{systembox}

\begin{userbox}
\scriptsize
\textbf{Background.} \textcolor{blue}{\{Scenario\}}

\smallskip
\textbf{Negotiable topics.} \textcolor{blue}{\{Topics\}}

\smallskip
\textbf{Conversation} (``Mediator'' refers to you)\textbf{.} \textcolor{blue}{\{Conversation Log\}}

\smallskip
Respond only with JSON:
\begin{verbatim}
{ "thought": "reasoning about whether and why to intervene", "should_engage": true/false }
\end{verbatim}
\end{userbox}
\end{promptbox}
\vspace{-0.25cm}
\caption{Prompt for mediator intervention decision.}
\vspace{-0.3cm}
\label{tab:prompt-mwhen}
\end{table*}

\begin{table*}[p]
\centering
\begin{promptbox}{Mediator: how-to-intervene generation}
\begin{systembox}
\scriptsize
You are the Mediator in a multi-party negotiation. Your role is to craft an intervention for the participants, based on the conversation history and scenario context.

\smallskip
\textbf{Task.}
\begin{itemize}[leftmargin=*, noitemsep, topsep=2pt]
\item Send a helpful and concise utterance that assists the participants or moves the discussion forward.
\item Stay sensitive to the social dynamics of the conversation and the participants' sentiments.
\item Be proactive in mediating the conversation and offering helpful guidance.
\item Use plain language, with no turn labels, headers, option labels, or bullet points.
\end{itemize}
\end{systembox}

\begin{userbox}
\scriptsize
\textbf{Background.} \textcolor{blue}{\{Scenario\}}

\smallskip
\textbf{Negotiable topics.} \textcolor{blue}{\{Topics\}}

\smallskip
\textbf{Conversation} (``Mediator'' refers to you)\textbf{.} \textcolor{blue}{\{Conversation Log\}}

\smallskip
\textbf{Previous thought.} \textcolor{blue}{\{When-to-intervene Thought\}}

\smallskip
Respond only with JSON:
\begin{verbatim}
{ "thought": "reasoning about what to say and why", "utterance": "your spoken utterance here" }
\end{verbatim}
\end{userbox}
\end{promptbox}
\vspace{-0.25cm}
\caption{Prompt for mediator intervention generation.}
\vspace{-0.3cm}
\label{tab:prompt-mhow}
\end{table*}
\clearpage

\begin{table*}[p]
\centering
\begin{promptbox}{Topic-localized Evaluator}
\begin{systembox}
\scriptsize
You are an expert in negotiation analysis. You analyze a full conversation and score the agreement between two parties on a specific topic.

\smallskip
\textbf{Task.} Read the full conversation. Identify every turn where either party meaningfully discusses or shifts its position on this topic. For each such turn, record (1) the agreement score (1--5) based on everything said up to that turn, and (2) each party's current stance, expressed with the option label(s) above (e.g.\ ``(A)'', ``(B)''), or a brief description if it matches no option. Omit turns that do not address the topic; scores carry forward from the last relevant turn.
\end{systembox}

\begin{userbox}
\scriptsize
\textbf{Background.} \textcolor{blue}{\{Scenario\}}

\smallskip
\textbf{Topic to analyze.} \textcolor{blue}{\{Topic\}}

\smallskip
\textbf{Parties.} \textcolor{blue}{\{Parties\}}

\smallskip
\textbf{Conversation.} \textcolor{blue}{\{Conversation Log\}}

\smallskip
Return JSON in exactly this format:
\begin{verbatim}
{
  "relevant_turns": [3, 7, 12],
  "agreement_score": [
    {"turn_id": 3, "reason": "brief reason",
     "score": int,
     "party_stances": {
       "party_A": "short description for stance",
       "party_B": "short description for stance"}},
    {"turn_id": 7, "reason": "brief reason",
     "score": int,
     "party_stances": {
       "party_A": "(B)", "party_B": "(C)"}}
  ]
}
\end{verbatim}
\end{userbox}
\end{promptbox}
\vspace{-0.25cm}
\caption{Prompt for topic-localized evaluation.}
\vspace{-0.3cm}
\label{tab:prompt-eval}
\end{table*}

\clearpage

\clearpage

\end{document}